%% file: acl.tex
\pdfoutput=1
\documentclass[11pt]{article}

\usepackage[]{acl}

\usepackage{times}
\usepackage{latexsym}
\usepackage{multirow}
\usepackage{url}
\usepackage{tabularx}
\usepackage{latexsym}
\usepackage{balance}
\usepackage{bm}
\usepackage{alltt}
\usepackage{algorithm}
\usepackage{algorithmic}
\usepackage{amssymb}
\usepackage{amsmath}
\usepackage{url}
\usepackage{makecell}
\usepackage{multirow}
\usepackage{subfig}
\usepackage{graphicx} 
\usepackage{color}
\usepackage{colortbl}
\usepackage{textcomp}
\usepackage{CJK}
\usepackage{enumitem}
\usepackage{booktabs}
\usepackage{microtype}
\usepackage{arydshln}
\usepackage{amsfonts}
\usepackage{wasysym}
\usepackage{eufrak}
\usepackage{pifont}
\usepackage[normalem]{ulem}
\useunder{\uline}{\ul}{}
\usepackage[T1]{fontenc}
\usepackage[utf8]{inputenc}

\usepackage{microtype}
\usepackage{pifont}
\usepackage{inconsolata}

\DeclareMathOperator*{\argmin}{arg\,min}
\newcommand{\seqlen}[1]{|#1|}

\newcommand{\given}{\,|\,}
\newcommand{\context}{x, y_{<t}}

\title{\textit{Learning from Imperfect Data}: Towards Efficient Knowledge Distillation of Autoregressive Language Models for Text-to-SQL}

\author{%
  Qihuang~Zhong$^{1}$,
  Kunfeng~Chen$^{1}$,
  Liang~Ding$^{2}$,
  \textbf{Juhua~Liu}$^{1}$\thanks{~~Corresponding Authors: Juhua Liu (e-mail: liujuhua@whu.edu.cn), Bo Du (e-mail: dubo@whu.edu.cn)},
  \textbf{Bo~Du}$^{1*}$,
  \textbf{Dacheng~Tao}$^{3}$ \\
  \fontsize{9.0pt}{\baselineskip}\selectfont $^{1}$ School of Computer Science, National Engineering Research Center for Multimedia Software, Institute of Artificial Intelligence\\ 
  \fontsize{9.0pt}{\baselineskip}\selectfont  and Hubei Key Laboratory of Multimedia and Network Communication Engineering, Wuhan University, China \\
  \fontsize{9.0pt}{\baselineskip}\selectfont $^{2}$ The University of Sydney, Australia \quad $^{3}$ Nanyang Technological University, Singapore \\
   \fontsize{9.0pt}{\baselineskip}\selectfont \texttt{\{zhongqihuang, chenkunfeng, liujuhua, dubo\}@whu.edu.cn}, \texttt{\{liangding.liam, dacheng.tao\}@gmail.com}
}

\begin{document}
\maketitle

\input{Section/0_abstract}
\input{Section/1_introduction}
\input{Section/2_preliminary}
\input{Section/3_method}
\input{Section/4_experiments}
\input{Section/5_related_work}
\input{Section/6_conclusion}

\bibliography{emnlp2024}

\input{Section/7_appendix}

\end{document}

%% file: Section/0_abstract.tex
\begin{abstract}
Large Language Models (LLMs) have shown promising performance in text-to-SQL, which involves translating natural language questions into SQL queries. However, current text-to-SQL LLMs are computationally expensive and challenging to deploy in real-world applications, highlighting the importance of compressing them. To achieve this goal, knowledge distillation (KD) is a common approach, which aims to distill the larger teacher model into a smaller student model. While numerous KD methods for autoregressive LLMs have emerged recently, it is still under-explored whether they work well in complex text-to-SQL scenarios. To this end, we conduct a series of analyses and reveal that these KD methods generally fall short in balancing performance and efficiency. In response to this problem, we propose to improve the \textbf{K}D with \textbf{I}mperfect \textbf{D}ata, namely \texttt{KID}, which effectively boosts the performance without introducing much training budget. The core of \texttt{KID} is to efficiently mitigate the training-inference mismatch by simulating the cascading effect\footnote{The error at the early step will affect the future predictions during the autoregressive inference~\cite{agarwal2023gkd}.} of inference in the imperfect training data. Extensive experiments on 5 text-to-SQL benchmarks show that, \texttt{KID} can not only achieve consistent and significant performance gains (up to +5.83\% average score) across all model types and sizes, but also effectively improve the training efficiency. 
\end{abstract}

%% file: Section/1_introduction.tex
\section{Introduction}
\label{sec:intro}

Text-to-SQL, which aims to translate a user's natural language question into an executable and accurate SQL query, is a transformative application of large language models (LLMs)~\cite{katsogiannis2023survey,li2024codes,pourreza2024din}. However, with the scaling of model size, the inference and deployment of LLM-based text-to-SQL systems become more computationally expensive and memory intensive, hindering the development of real-world industrial applications that require low inference latency~\cite{sun2023exploratory}. Hence, it is crucial and green to compress these text-to-SQL LLMs and accelerate the inference, while not losing much performance~\cite{schwartz2020green,zhu2023survey}.

\begin{figure}[t]
    \centering
    \includegraphics[width=0.47\textwidth]{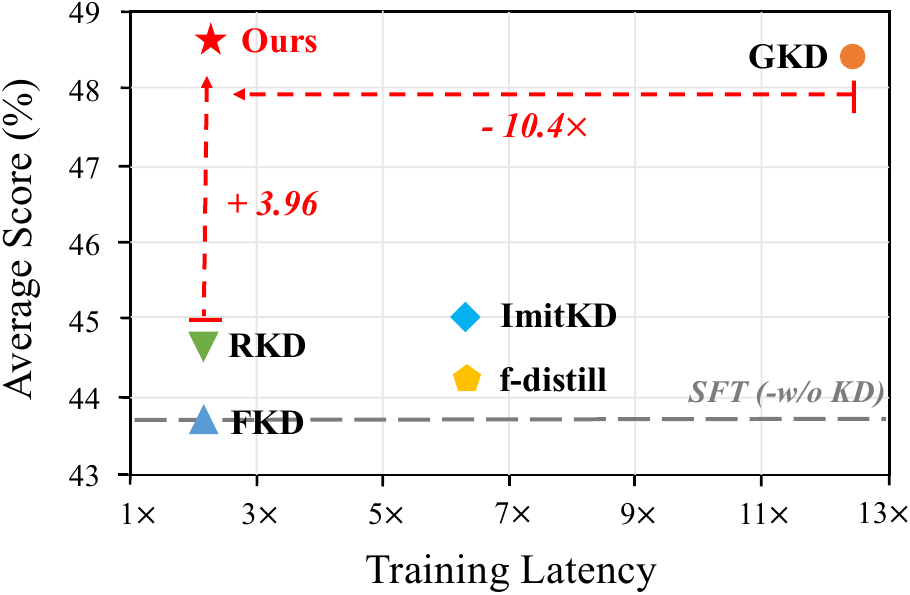}
    \caption{\textbf{Comparisons of different KD methods} for distilling the student model (QWen1.5-0.5B) from the teacher (QWen1.5-4B). The x-axis denotes the training latency relative to the SFT baseline, while the y-axis denotes the average performance of students on several popular text-to-SQL benchmarks. The evaluation details are in \S\ref{sec:experiments}. We see that our method achieves the best trade-off between performance and efficiency.}
    \label{fig:intro}
\end{figure}

A common model compression approach is knowledge distillation (KD), which involves compressing a large teacher model by distilling its knowledge into a small student model~\cite{hinton2015distilling, kim2016sequence}. Recently, numerous KD methods for autoregressive LLMs have emerged~\cite{gu2023knowledge,agarwal2023gkd,xu2024survey}, but most of them focus on the general instruction-tuning scenarios. Different from the general tasks that allow for flexible and diverse outputs, text-to-SQL is more challenging, as it requires the LLMs to precisely output the table/column name. Even a minor error in the SQL query could lead to the wrong result. Unfortunately, it is still under-explored whether these KD methods work well for text-to-SQL LLMs.

To this end, we conduct preliminary experiments by applying 5 representative KD methods to distill the QWen-family LLMs~\cite{bai2023qwen} on the popular text-to-SQL benchmark, \textit{i.e.}, Spider~\cite{yu2018spider}. We find that the performance gains of these KD methods mainly rely on the model-generated data, which is effective but hard to obtain. Specifically, although the model-generated data can alleviate the training-inference mismatch (\textit{i.e.}, difference between teacher-forcing training and autoregressive inference~\cite{pang2020text}) and achieves remarkable performance, it requires the student model to autoregressively generate in an online fashion, leading to unbearable training latency. As illustrated in Figure~\ref{fig:intro}, GKD~\cite{agarwal2023gkd} training with model-generated data performs well but greatly suffers from training inefficiency. Thus, there raises a question: \textit{whether we can mitigate the training-inference mismatch more efficiently}?

Motivated by this, we propose a simple-yet-effective approach to improve KD, namely \texttt{KID}, and achieve a better trade-off between performance and efficiency. The core of \texttt{KID} is to force the student to rewrite the ground-truth training data into imperfect one, and then learn how to calibrate these imperfect data. 
% Specifically, instead of autoregressively generating the on-policy data, the generation processes of imperfect data only require one-pass forward, which is more efficient.
Intuitively, by introducing some errors in the imperfect data, we can simulate the cascading effect of inference during training processes, thus mitigating the training-inference mismatch. More specifically, instead of autoregressively generating the on-policy data, the generation processes of imperfect data only require one-pass forward, which is more efficient and affordable.
% the imperfect data can simulate the cascading effect\footnote{The error at early step will affect the future predictions~\cite{agarwal2023gkd}.} of inference. That is, training with these imperfect data can mitigate the training-inference mismatch. 
Moreover, by doing so, we can also encourage the student to learn how to calibrate these imperfect tokens and further improve the KD performance.

We evaluate \texttt{KID} on a variety of popular text-to-SQL benchmarks, including BIRD~\cite{li2024can}, Spider~\cite{yu2018spider} and its variants, upon 3 types of autoregressive LLMs: QWen~\cite{bai2023qwen}, CodeGen~\cite{nijkamp2022codegen} and LLaMA~\cite{touvron2023llamav2}. Results show that \texttt{KID} can not only achieve a better trade-off between performance and efficiency, but also bring consistent and significant improvements (up to +5.83\% average score) among all model types and sizes. Moreover, compared to the standard KD, \texttt{KID} can effectively improve the robustness of students.

\paragraph{Contributions.} Our main contributions are:
\begin{itemize}
    \item We reveal that current KD methods for text-to-SQL LLMs generally fall short in balancing performance and efficiency.
    \item We propose a simple-yet-effective approach (\texttt{KID}) to effectively improve KD performance without introducing much training budget.
    \item Extensive experiments show that \texttt{KID} outperforms the standard KD by a large margin and effectively improves the student's robustness.
\end{itemize}

%% file: Section/2_preliminary.tex
\section{Preliminary}
\label{sec:preliminaries}
\subsection{Task Formulation}
\label{sec:2.1}
Text-to-SQL aims to convert a natural language question $\mathcal{Q}$ into a SQL query $\mathcal{Y}$, which is executable and can accurately retrieve relevant data from a database $\mathcal{D}$. The database $\mathcal{D}$ usually contains the schema (\textit{i.e.}, tables and columns) and metadata, containing column types/values, primary keys, foreign key relations and \textit{etc}~\cite{zhong2017seq2sql}. 
% In the context of LLM-based Text-to-SQL, LLMs are used to directly generate the query $\mathcal{Y}$. 
Specifically, given an LLM $\mathcal{M}$ and a prompt template $\mathcal{P}$, we enforce the $\mathcal{M}$ to autoregressively generate an output sequence $\mathcal{Y}$ conditioned on the $\mathcal{P}(\mathcal{Q}, \mathcal{D})$, which can be formulated as:
% \begin{align*}
\begin{equation}
    \mathcal{Y}_t \sim \mathbb{P}_\mathcal{M}(\mathcal{Y}_t \mid \mathcal{P}(\mathcal{Q},\mathcal{D}), \mathcal{Y}_{<t}),
% \end{align*}
\end{equation}
where $\mathbb{P}_\mathcal{M}(\mathcal{Y}_t \mid \mathcal{P}(\mathcal{Q},\mathcal{D}), \mathcal{Y}_{<t})$ is the probability for the next token, and $\mathcal{Y}_t$ is the $t$-th token of $\mathcal{Y}$.

\subsection{Knowledge Distillation of LLMs}
Knowledge Distillation (KD) aims to compress a large teacher model $\mathcal{M}_p$ by distilling its knowledge into a small student model $\mathcal{M}_q^\theta$ parameterized by $\theta$. Given a divergence function $\mathcal{F}$ and a training set $\mathcal{G}$, we can train the student model as follows:
\begin{equation}
    \theta^{*} := \argmin \mathbb{E}_{(x,y)\sim \mathcal{G}}[{\mathcal{F}(\mathcal{M}_q \| \mathcal{M}_q^\theta) (y|x)}],
  \label{eq:distill_loss}
\end{equation}
where $(x,y)$ is the task-specific input-output pair\footnote{For text-to-SQL task in \S\ref{sec:2.1}, $x$ refers to the input question $\mathcal{P}(\mathcal{Q}, \mathcal{D})$ and $y$ refers to the output SQL query $\mathcal{Y}$.} of $\mathcal{G}$, and ${\mathcal{F}(\mathcal{M}_q \| \mathcal{M}_q^\theta) (y|x)} = \frac{1}{\seqlen{y}} \sum_{t=1}^{|y|} \mathcal{F}\big(p(\,\cdot \given \context) \| q^\theta(\,\cdot \given \context) \big)$ is the divergence between the teacher and student distributions, denoted as $p$ and $q^\theta$, respectively. 
% For short, we simply denote the distributions of teacher and student as $p$ and $q^\theta$, respectively.
The choices of training set $\mathcal{G}$ and divergence function $\mathcal{F}$ give rise to different possible KD algorithms, \textit{e.g.}, Forward KD (FKD)~\cite{hinton2015distilling}, Reverse KD (RKD)~\cite{gu2023knowledge}, f-distill~\cite{wen2023f}, ImitKD~\cite{lin2020autoregressive} and GKD~\cite{agarwal2023gkd}. The summary of these representative KD algorithms is shown in Table~\ref{tab:summary}. 

The common divergences for KD contain the Forward Kullback-Leibler (FKL)~\cite{van2014renyi}, Reverse KL (RKL)~\cite{malinin2019reverse}, Jensen–Shannon divergence (JSD)~\cite{fuglede2004jensen} and total variation distance (TVD)~\cite{verdu2014total}. The details of these divergences can be found in Appendix~\ref{appendix:divergence_fn}. On the other hand, $\mathcal{G}$ may consist of input-output pairs in the original training set (denoted as \textbf{ground-truth dataset}), or sequences generated from teacher $\mathcal{M}_p$ or student $\mathcal{M}_q^\theta$ (denoted as \textbf{model-generated dataset}). For the data generated by $\mathcal{M}_p$, we feed the input into the $\mathcal{M}_p$ and obtain the teacher's output beforehand and keep them fixed during training. Conversely, for the data generated by $\mathcal{M}_q^\theta$, since the student is continuously updated, we obtain the student's output in an online fashion. Such online generated data is also called ``on-policy data'' by~\citet{agarwal2023gkd}.

\begin{table}[]
\centering
\scalebox{0.71}{
\begin{tabular}{lll}
\toprule
\textbf{Method} & \textbf{Divergence} & \textbf{Training Dataset} \\ \midrule \midrule
\multicolumn{3}{l}{\textit{Data type: Fixed dataset}} \\ \hdashline
\textbf{FKD} & FKL & Ground-truth data \\
\textbf{RKD} & RKL & Ground-truth data \\ \midrule
\multicolumn{3}{l}{\textit{Data type: Model-generated dataset}} \\ \hdashline
\textbf{f-distill} & TVD & Data generated by $\mathcal{M}_p$ and $\mathcal{M}_q^\theta$ \\
\textbf{ImitKD} & FKL & Ground-truth+data generated by $\mathcal{M}_q^\theta$ \\
\textbf{GKD} & FKL/RKL/JSD & On-policy data generated by $\mathcal{M}_q^\theta$ \\
\midrule
\textbf{\texttt{KID}} & RKL & Imperfect ground-truth data \\
\bottomrule
\end{tabular}
}
\caption{\textbf{Summary of various KD algorithms in terms of training data and divergence}. Notably, $\mathcal{M}_p$ and $\mathcal{M}_q^\theta$ denote the teacher and student models, respectively.}
\label{tab:summary}
\end{table}

\subsection{Empirical Analyses}
% To the best of our knowledge, there is a lack of an exploratory study on KD for autoregressive LLMs in the field of text-to-SQL. Hence, we attempt to explore whether the aforementioned KD algorithms work well for text-to-SQL LLMs in this part.
As mentioned in \S\ref{sec:intro}, it is under-explored whether the aforementioned KD algorithms work well for text-to-SQL LLMs. Hence, we conduct preliminary experiments to investigate it in this part. 

\paragraph{Setting.} We conduct experiments by first fine-tuning larger LLMs on the original training dataset as teachers. Then, we use different KD methods to distill a smaller student with the teacher’s guidance. Here, we use the QWen1.5-0.5B~\cite{bai2023qwen} as the student and use the other QWen-family models (\textit{i.e.}, QWen1.5-1.8B/-4B/-7B) as teachers. Spider~\cite{yu2018spider} is used as training data, and the models are evaluated on the development set. We follow~\citet{li2024codes} and use the ``Execution Accuracy'' as metric to quantify the model output.

\begin{table}[]
\setlength{\tabcolsep}{11pt}
\scalebox{0.8}{
\begin{tabular}{lcccc}
\toprule
Method &Divergence & \multicolumn{1}{l}{1.8B} & \multicolumn{1}{l}{4B} & \multicolumn{1}{l}{7B} \\ \midrule \midrule
\multicolumn{5}{l}{\textit{Training data: \textbf{Fixed dataset}}} \\ \hdashline
FKD&FKL & 57.3 & 57.4 & 57.3 \\
 RKD&RKL & 62.7 & 60.1 & 61.5 \\ \midrule
\multicolumn{5}{l}{\textit{Training data: \textbf{Model-generated dataset}}} \\ \hdashline
f-distill &TVD & 57.6 & 58.6 & 59.6 \\
ImitKD &FKL & 58.3 & 59.5 & 59.1 \\
GKD-FKL &FKL & 61.1 & 62.1 & 60.7 \\
GKD-RKL & RKL & \textbf{62.9} & \textbf{63.8} & \textbf{64.3} \\
GKD-JSD & JSD & 62.8 & 62.7 & \textbf{64.3} \\
\bottomrule
\end{tabular}
}
\caption{\textbf{Preliminary experimental results (\%) of various KD methods}. We report the execution accuracy of QWen1.5-0.5B distilling from QWen1.5-\{1.8B, 4B, 7B\} on the Spider benchmark. Best results are in \textbf{bold}.}
\label{tab:pre_result}
\end{table}

\begin{figure}[t]
    \centering
    \includegraphics[width=0.42\textwidth]{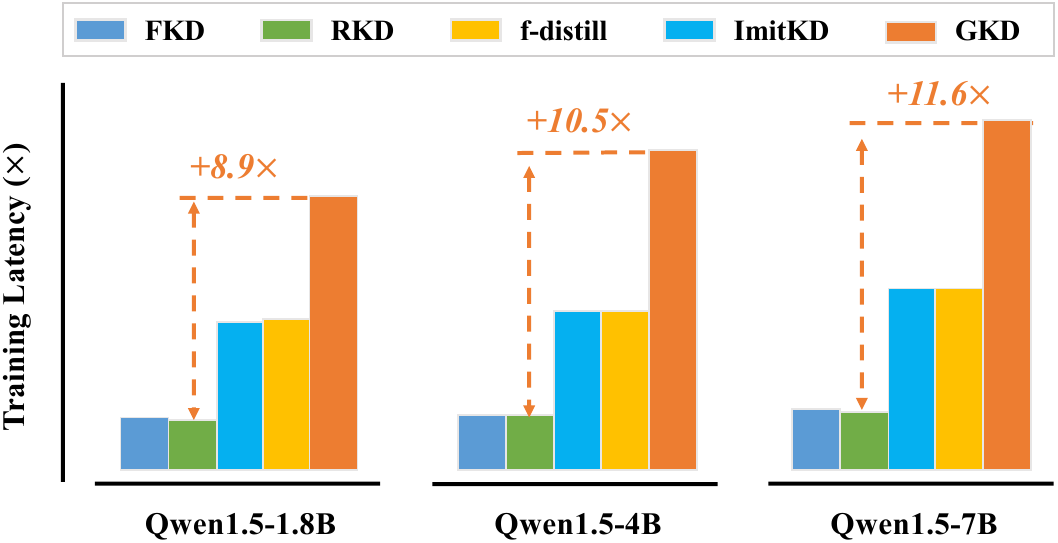}
    \caption{\textbf{Comparisons of training latency between various KD methods}. The x-axis denotes the teacher models, and the y-axis denotes the training latency relative to the SFT baseline. For ease of illustration, we only report the results of RKL divergence for GKD.}
    \label{fig:pre_latency}
\end{figure}

\begin{figure*}[t]
    \centering
    \includegraphics[width=0.95\textwidth]{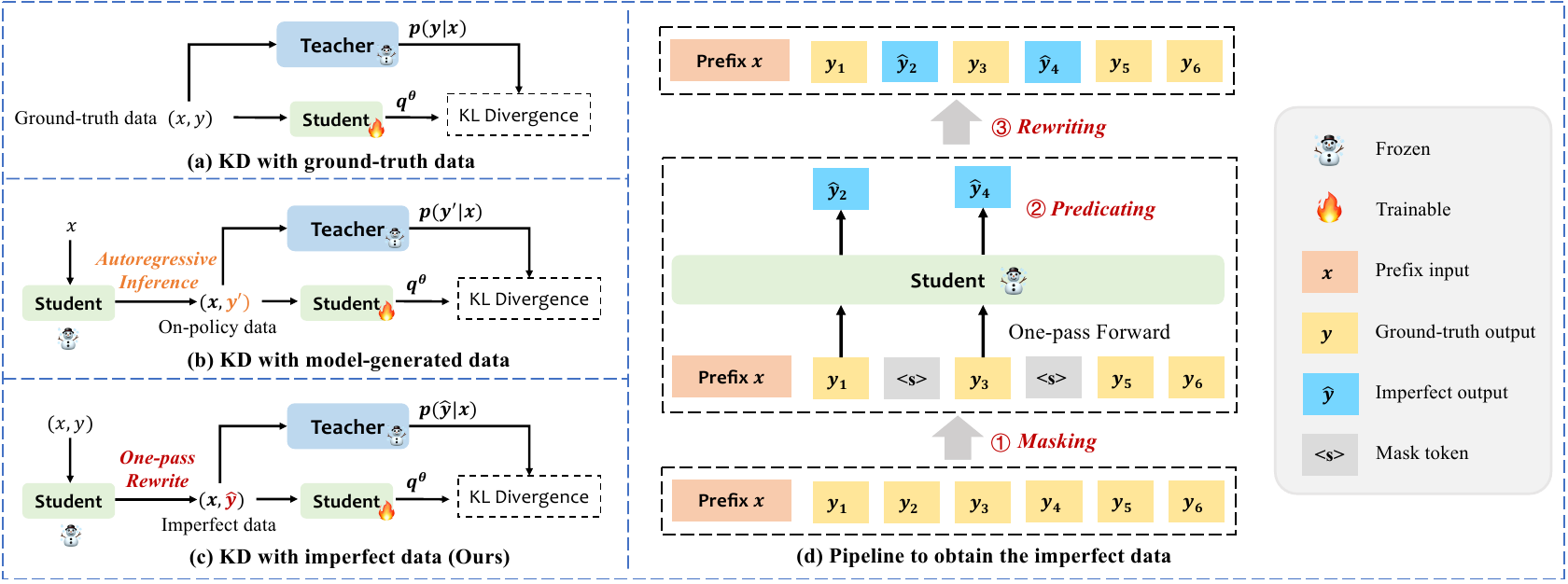}
    \caption{\textbf{Illustrations of different KD methods}: \textbf{(a)} KD methods with ground-truth data, \textbf{(b)} KD methods with model-generated data and \textbf{(c)} our \texttt{KID} method with imperfect data. Additionally, we show \textbf{(d)} the pipeline to obtain the imperfect data, which contains three-stage processes: \ding{182} \textbf{\textit{masking}}, \ding{183} \textbf{\textit{predicting}} and \ding{184} \textbf{\textit{rewriting}}.}
    \label{fig:method}
\end{figure*}

\paragraph{Findings.} The comparative results are listed in Table~\ref{tab:pre_result},
% Moreover, we also illustrate the training latency of various KD methods in Figure~\ref{fig:pre_latency}. 
from which we empirically find that: 

\paragraph{Reverse KL is more suitable for distilling the text-to-SQL LLMs.} We first analyze the impact of different divergence functions, and find that RKL generally outperforms the other divergences, \textit{e.g.}, FKD (57.4\%) \textit{v.s.} RKD (60.1\%) and GKD-FKL (62.1\%) \textit{v.s.} GKD-RKL (63.8\%). This is similar to the statements of prior studies~\cite{gu2023knowledge,wu2024rethinking}, as they argue that Reverse KL shows mode-seeking behaviors, \textit{i.e.}, it does not force the student to fit all teacher's distributions, but assigns high probabilities to teacher’s large modes and ignores the small ones. In the context of text-to-SQL, the output tokens (\textit{e.g.}, table/column name and value) are usually precise and low-diversity, and enforcing the student to learn the high-probability regions could lead to better performance.

\paragraph{Model-generated datasets perform better but suffer from training inefficiency.} By comparing the KD results between ground-truth datasets and model-generated datasets, we find that model-generated datasets perform better than the fixed ground-truth ones, especially the on-policy dataset generated by students (\textit{i.e.}, GKD). This is because that student-generated dataset can alleviate the training-inference mismatch, \textit{i.e.}, the discrepancy between teacher-forcing training and free-run inference. Despite its remarkable performance, it requires the student to autoregressively generate the output in an online manner, which will lead to unaffordable training latency. This can be empirically proven by the results in Figure~\ref{fig:pre_latency}, as the training latency of GKD is much higher than those trained on ground-truth datasets.

% Although those methods trained with model-generated dataset perform well, they lead to unaffordable training expenses. Motivated by this, we 

%% file: Section/3_method.tex
\section{Improving Knowledge Distillation with Imperfect Data}
\label{sec:method}

\paragraph{Motivation and Overview.} 
Based on the observation in \S\ref{sec:preliminaries}, we recognize that 
% current KD methods generally fall short in balancing the performance and efficiency. While those relying on model-generated dataset can alleviate the training-inference mismatch and achieve better performance, they greatly suffer from training inefficiency.
the key for improving the performance KD is to alleviate the training-inference mismatch. However, the current KD methods relying on model-generated datasets usually suffer from training inefficiency, \textit{i.e.}, they fail to balance the performance and efficiency. 
Thus, there raises a question: \textit{whether we can mitigate the training-inference mismatch more efficiently}? Motivated by this, we propose to improve KD with imperfect data (\texttt{KID}), which effectively and efficiently boosts the performance by simulating the cascading effect of inference during training. 
% The core of \texttt{KID} is to enforce the student to first rewrite the ground-truth data into imperfect data, which is more closer to inference scenarios, and then learn how to calibrate these imperfect data. 
The illustration of \texttt{KID} is shown in Figure~\ref{fig:method}.

\paragraph{Intuition of \texttt{KID}.}
% As stated by prior studies~\cite{pang2020text,gu2023knowledge,agarwal2023gkd}, the training-inference mismatch mainly comes from the difference between teacher-forcing mode (training) and autoregressive generation mode (inference). Specifically, during training, the LLM conditions on the ground-truth tokens. However, during inference, it conditions on the model-generated tokens and has a cascading effect where error at early step will affect the future predictions. 
As stated by prior studies~\cite{pang2020text,agarwal2023gkd}, the training-inference mismatch mainly comes from the cascading effect of inference. Specifically, during training, LLMs condition on ground-truth tokens. However, during inference, they condition on the model-generated tokens, which might be wrong and affect the future predictions.
Intuitively, enforcing the student to rewrite the ground-truth training data into imperfect one, \textit{i.e.}, introducing some errors during training, can simulate the cascading effect of inference and thus mitigate the training-inference mismatch. Moreover, by encouraging the student to learn how to calibrate these imperfect tokens, \texttt{KID} can further improve the performance.
% we can also encourage the student to learn how to calibrate these imperfect tokens and further improve the performance.

\paragraph{Pipeline to Obtain the Imperfect Data.}
The key technique of \texttt{KID} is to rewrite the ground-truth data into an imperfect one. Specifically, the generation of imperfect data consists of three-stage processes: \ding{182} \textbf{\textit{masking}}, \ding{183} \textbf{\textit{predicting}} and \ding{184} \textbf{\textit{rewriting}}. In practice, we \ding{182} first sample $\alpha$ of tokens\footnote{The analysis of sampling ratio $\alpha$ can be found in~\S\ref{sec:ablation}.} from the ground-truth output $y$ and mask them with a special token (\textit{e.g.}, ``\texttt{<s>}''). For sampling the tokens, we design some strategies: 1) ``Random'': randomly sampling, 2) ``Uniform'': uniformly sampling, 3) ``Hard'': sampling $\alpha$ of tokens with the lowest confidence; 4) ``Easy'': sampling $\alpha$ of tokens with the highest confidence. More specifically, for 3) and 4), we feed the original sequence $y$ into the student for obtaining prediction probabilities $q^\theta_i$, and then compute the entropy of $q^\theta_i$ as the confidence\footnote{Intuitively, the tokens with high entropy value are hard-to-learn, as the model predict them with low confidence towards the gold labels~\cite{zhong2023self}.}.

After masking the spans of $y$, we \ding{183} then generate imperfect tokens to fill in the spans. Specifically, we feed the masked sequence into the student to generate predictions with a one-pass forward process. Finally, given the predicted imperfect tokens on the masking place, we \ding{184} rewrite the ground-truth $y$ into the imperfect one $\hat{y}$. 
% Note that we perform the above processes in an online fashion.

\paragraph{Training of \texttt{KID}.}
During training, given a mini-batch of input-output pairs $(x,y)$, we first perform the above processes to obtain the imperfect data $(x,\hat{y})$. Then, we can train the student model with the teacher's guidance. As shown in \S\ref{sec:preliminaries}, Reverse KL is more suitable for text-to-SQL task, and we thus use it as the divergence function in our \texttt{KID}. 
Moreover, since our \texttt{KID} require sampling from a student, which may generate poor samples at the beginning of training and make the distilling more difficult, we follow prior works~\cite{wen2023f,gu2023knowledge} and combine the KD loss in Eq.~\ref{eq:distill_loss} with an auxiliary maximum likelihood estimation (MLE) loss. Specifically, the MLE loss enforces the student to predict the ground-truth target sequences $y$. Notably, for a fair comparison, we also add the auxiliary MLE loss into the baseline KD methods that rely on the ground-truth data.

% The illustration of \texttt{KID} is shown in Figure~\ref{}.

% Based on the observation in \S\ref{sec:preliminaries}, we recognize that there is a lack of KD method that achieves better trade-off between the performance and efficiency. Although those trained with model-generated data performs well, they lead to unaffordable training expenses. Thus, there raises a question: \textit{whether we can propose an efficient KD method that performs well without using model-generated data}? Motivated by this, we propose to improve KD with imperfect data (\texttt{KID}). The core of \texttt{KID} is to mitigate the training-inference mismatch by rewriting the ground-truth data into imperfect data, which is more closer to inference scenarios. The illustration of \texttt{KID} is shown in Figure~\ref{}.

%% file: Section/4_experiments.tex
\input{Tables/main_results}

\section{Experiments}
\label{sec:experiments}
\subsection{Setup}
\paragraph{Tasks and Datasets.}
We conduct our main experiments on two popular text-to-SQL benchmarks, \textit{i.e.}, Spider~\cite{yu2018spider} and BIRD~\cite{li2024can}. For each task, models are trained with the original training set and evaluated on the development set, denoted as Spider-dev and BIRD-dev, respectively. Moreover, following prior studies~\cite{li2023resdsql,li2024codes}, we also evaluate the models trained with the Spider dataset on three more challenging robustness benchmarks, \textit{i.e.}, Spider-DK~\cite{gan2021exploring}, Spider-Realistic~\cite{deng2021structure} and Spider-Syn~\cite{gan2021towards}. 

For evaluation on Spider-family benchmarks, we utilize two widely-used metrics, \textit{i.e.}, ``Execution Accuracy'' (EX)~\cite{yu2018spider} and ``Test-Suite Accuracy'' (TS)~\cite{zhong2020semantic}. For BIRD, we simply use the EX as the evaluation metric. Notably, BIRD offers external knowledge for guiding the generation of SQL queries. Considering that such external knowledge is usually unavailable in the real world, we follow~\citet{li2024codes} and perform the evaluation in two settings: without (``w/o EK'') and with (``w/ EK'') external knowledge. The details of all tasks are shown in Appendix~\ref{appendix_data}.

\paragraph{Models.} We evaluate \texttt{KID} on three types of LLMs with various sizes: QWen1.5~\cite{bai2023qwen} (\textit{student}: 0.5B, \textit{teachers}: 1.8B, 4B, 7B), CodeGen~\cite{nijkamp2022codegen} (\textit{student}: 350M, \textit{teachers}: 2B), and LLaMA2 (\textit{student}: TinyLLaMA-1.1B~\cite{zhang2024tinyllama}, \textit{teachers}: 7B~\cite{touvron2023llamav2}). 
% We train each model with a batch size of 16 and a peak learning rate of 2e-4. 
% \footnote{Since there are no existing official LLaMA smaller than 7B, we use the other re-produced smaller TinyLLaMA-1.1B from~\citet{zhang2024tinyllama} as the student.}
All models are trained with a popular parameter-efficient fine-tuning method, \textit{i.e.}, LoRA~\cite{hu2021lora}. The details of all training hyper-parameters can be found in Appendix~\ref{appendix_training}.

\paragraph{Baselines.} We consider 5 cutting-edge KD baselines in our main experiment: Forward KD (FKD)~\cite{hinton2015distilling}, Reverse KD (RKD)~\cite{gu2023knowledge}, f-distill~\cite{wen2023f}, ImitKD~\cite{lin2020autoregressive} and GKD\footnote{As shown in Table~\ref{tab:pre_result}, GKD with RKL divergence (\textit{i.e.}, GKD-RKL) performs best, and we thus only report the results of GKD-RKL for GKD in the following content.}~\cite{agarwal2023gkd}. For reference, we also report the performance of teachers as the upper bound. We use the codebase of~\citet{liu2023online} to implement these baselines and distill students.

\subsection{Main Results}
\paragraph{\texttt{KID} achieves a better trade-off between the KD performance and efficiency.} The main results on QWen-family models are listed in Table~\ref{tab:main_qwen}. As seen, most KD methods outperform the SFT baseline, while introducing extra training budgets. Training with the on-policy data, GKD achieves much better performance than the other counterparts. However, the computational budget of GKD is not affordable, as it leads to up to 13.9$\times$ training latency against the SFT baseline. Conversely, our \texttt{KID} can not only achieve comparable or even better performance than GKD, but also effectively reduce the training latency. These results can prove the superiority of our method.

\input{Tables/main_results2}

\begin{figure}[t]
\centering
    \includegraphics[width=0.45\textwidth]{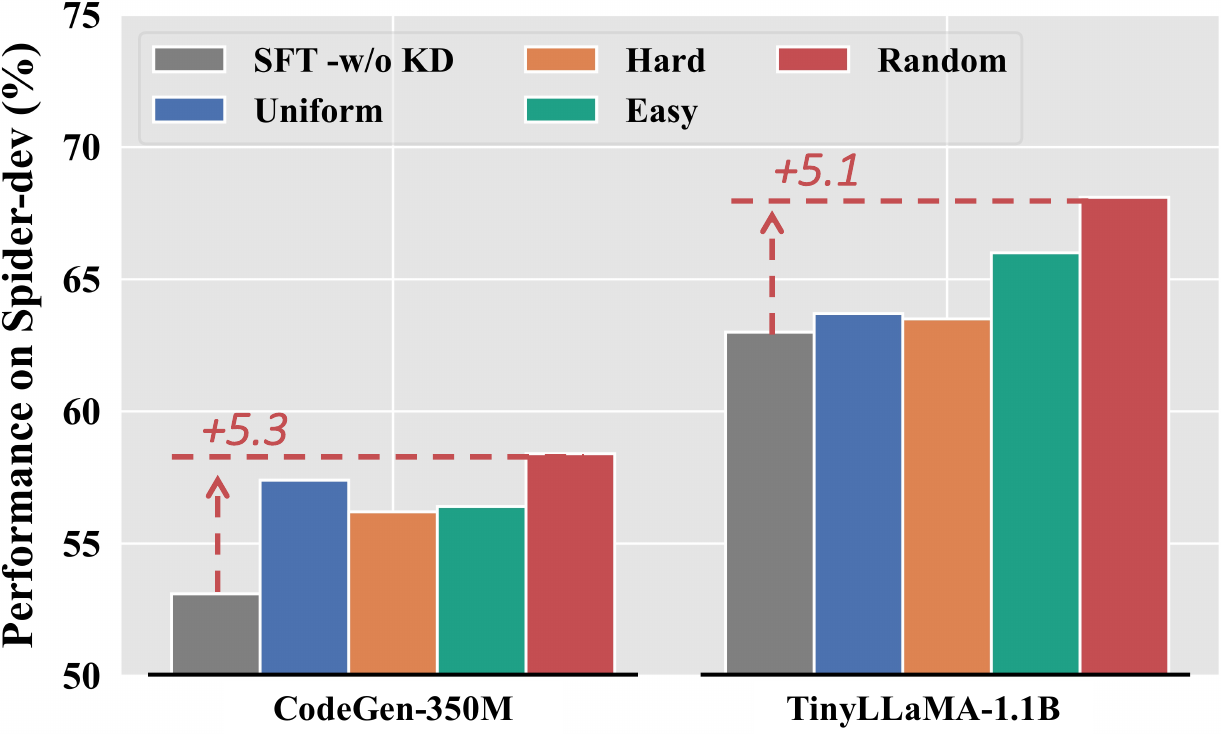}
    \caption{\textbf{Analysis of different masking strategies}. The y-axis denotes the EX performance on Spider-dev. For reference, we also report the results of SFT.}
    \label{fig:ablation_strategies}
\end{figure}

\paragraph{\texttt{KID} brings consistent and significant performance gains among all model sizes and types.} In addition to QWen-family models, we also apply our method on CodeGen and LLaMA models, and report the results in Table~\ref{tab:main_llama}. Notably, due to the space limitation, we only report the contrastive results of two most relevant KD counterparts, \textit{i.e.}, RKD and GKD. From the results of Table~\ref{tab:main_qwen} and~\ref{tab:main_llama}, it can be found that our \texttt{KID} consistently outperforms the other KD counterparts and brings significant performance gains (up to +5.83\% average score) against the SFT baseline among all model sizes and types, indicating its universality. 

\paragraph{\texttt{KID} effectively improves the robustness of distilled models.} Spider-DK, Spider-Syn, and Spider-Realistic are widely-used challenging benchmarks to investigate the robustness of text-to-SQL models. Contrastive results on these benchmarks show that our \texttt{KID} exhibits exceptional performance and effectively improves the robustness of distilled students. For example, when distilling CodeGen models, \texttt{KID} achieves gains of 2.7\% on Spider-DK (43.7\% to 46.4\%) and 2.1\% on Spider-Realistic (45.5\% to 47.6\%), comparing with the best counterpart. 

\subsection{Analysis of \texttt{KID}}
\label{sec:ablation}
We evaluate the impact of each component of our \texttt{KID}, including 1) masking strategies, 2) masking ratio $\alpha$, and 3) rewriting approach for obtaining the imperfect data. Additionally, we 4) perform the in-depth analysis on the training efficiency of \texttt{KID}.

\paragraph{Effect of different masking strategies.}
As mentioned in \S\ref{sec:method}, we introduce several strategies to select the tokens for masking. Here, we conduct experiments to analyze the impact of different masking strategies. Results of CodeGen-350M and TinyLLaMA-1.1B in Figure~\ref{fig:ablation_strategies} show that: 1) Our \texttt{KID} with various masking strategies consistently outperforms the SFT baseline. 2) Performance of difficulty-driven strategies (\textit{i.e.}, ``Easy'' and ``Hard'') is unstable, as paying too much attention to the easy-to-learn/hard-to-learn tokens might affect the learning of the other tokens and thus leads to sub-optimal performance. 3) The ``Random'' strategy achieves consistently better performance. We conjecture that such a random masking strategy is closer to the errors that are prone to occur during inference, as a model might predict incorrect tokens at any inference step. Thus, we use the ``Random'' strategy as our default setting.

\begin{figure}[t]
\centering
    \includegraphics[width=0.45\textwidth]{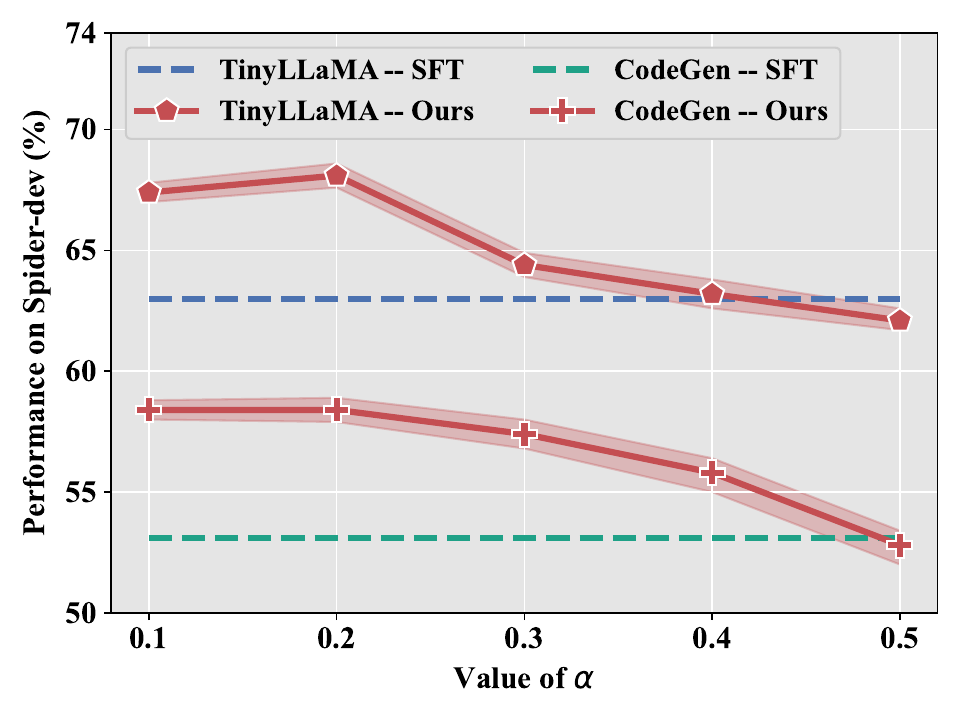}
    \caption{\textbf{Parameter analysis of masking ratio $\alpha$}. We report the EX results of TinyLLaMA-1.1B and CodeGen-350M on the Spider-dev.}
    \label{fig:ablation_ratio}
\end{figure}

\paragraph{Parameter analysis on $\alpha$.}
The $\alpha$ used to control the ratio of masking tokens is an important hyper-parameter. Here, we analyze its influence by evaluating the performance of \texttt{KID} with different $\alpha$, spanning \{0.1, 0.2, 0.3, 0.4, 0.5\} on Spider-dev. Figure~\ref{fig:ablation_ratio} illustrates the contrastive results. Compared with the SFT baseline, our \texttt{KID} consistently brings improvements across a certain range of $\alpha$ (\textit{i.e.}, 0.1 to 0.3), basically indicating that the performance of \texttt{KID} is not sensitive to $\alpha$. 2) Too large $\alpha$ values (\textit{e.g.}, 0.5) lead to performance degradation, as too many rewriting tokens might distort the sequence meaning and are challenging for models to calibrate. More specifically, the case of $\alpha = 0.2$ performs best, and we use this setting as default.

\begin{table}[]
\centering
\scalebox{0.82}{
\begin{tabular}{lcc}
\toprule
Method & CodeGen & TinyLLaMA \\ \midrule \midrule
SFT & 53.1 & 63.0 \\ \midrule
Vanilla \texttt{KID} & 55.1 & 66.0 \\ \hdashline
\quad \textbf{-w/ Masking-only} & 55.8 (\textcolor[RGB]{0,176,80}{${\uparrow0.7}$}) & 66.5 (\textcolor[RGB]{0,176,80}{${\uparrow0.5}$}) \\ 
\quad \textbf{-w/ Rewriting (Ours)} & 58.4 (\textcolor[RGB]{0,176,80}{${\uparrow\textbf{3.3}}$}) & 68.1 (\textcolor[RGB]{0,176,80}{${\uparrow\textbf{2.1}}$}) \\ 
\bottomrule
\end{tabular}
}
\caption{\textbf{Impact of rewriting approach of \texttt{KID}}. Notably, ``Vanilla \texttt{KID}'' means that we do not train with the imperfect data in our \texttt{KID}, ``-w/ Masking-only'' denotes that we directly use the sequence with masking spans as final imperfect data during the training of \texttt{KID}, and ``-w/ Rewriting (Ours)'' refers to the full \texttt{KID}.}
\label{tab:ablation_rewrite}
\end{table}

\paragraph{Impact of rewriting approach.} 
In the stage~\ding{184} of pipeline for obtaining the imperfect data, we rewrite the ground-truth data with the predicted imperfect tokens. To verify its effectiveness, we compare it with a simple alternative, \textit{i.e.}, directly using the sequence with masking spans (output of stage~\ding{182}) as final imperfect data $\hat{y}$, denoted as ``-w/ masking-only''.
% A key technology in our \texttt{KID} is the sequence span rewriting, which uses the model to rewrite the masking spans into the imperfect text for simulating the inference errors. To verify its effectiveness, we compare it with a simple alternative, \textit{i.e.}, directly using the sequence with masking spans as the input for KD. 
Table~\ref{tab:ablation_rewrite} shows the contrastive results (EX results on Spider-dev), in which we see that 1) the alternative approach equipped with \texttt{KID} outperforms the SFT, showing the superiority of our \texttt{KID}, and importantly, 2) our rewriting approach could further improve the results by a large margin against the simple alternative, \textit{e.g.,} +3.3\% gains on CodeGen-350M, indicating its effectiveness.

\begin{figure}[t]
\centering
    \includegraphics[width=0.45\textwidth]{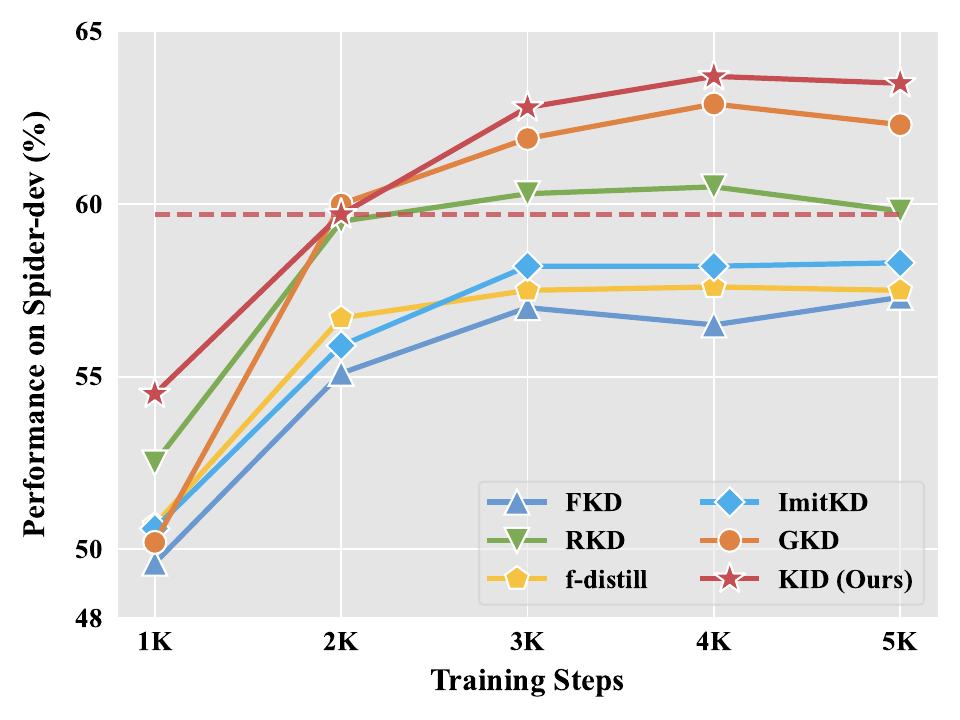}
    \caption{\textbf{Performance on Spider-dev of students (QWen1.5-0.5B) trained with different KD methods for the full training process}. QWen1.5-1.8B is used as the teacher. We see that \texttt{KID} achieves comparable performance with most counterparts at 2K training steps.}
    \label{fig:ablation_efficiency}
\end{figure}

\paragraph{Analysis of training efficiency.} In Table~\ref{tab:main_qwen}, we show that our \texttt{KID} effectively reduces the training latency compared to those counterparts based on model-generated data. Here, to further verify the training efficiency of \texttt{KID}, we present the performance of students trained with various KD methods across different training steps. QWen1.5-0.5B and 1.8B models are used as student and teacher, respectively. The results are illustrated in Figure~\ref{fig:ablation_efficiency}. As seen, \texttt{KID} can achieve comparable or even better performance than most KD counterparts with much fewer training steps, \textit{i.e.}, effectively improving the training efficiency. We attribute it to the higher data efficiency, since the imperfect data is closer to inference scenarios and can help the student better adapt to downstream generation.

\subsection{Discussion}

\paragraph{Does \texttt{KID} still work under larger model size gaps?} Here, to further prove the effectiveness of our \texttt{KID}, we attempt to apply it to distill the larger LLMs. In practice, we use our method to distill the Qwen1.5-32B teacher model into the Qwen1.5-0.5B student model, and report the contrastive results on Spider-family benchmarks in Table~\ref{tab:larger_teacher}. As seen, compared with the KD baselines, \texttt{KID} can still achieve much better performance among all benchmarks. \textit{These results indicate that our method can work well in the larger teacher models.}

\begin{table}[]
\centering
\scalebox{0.75}{
\begin{tabular}{lcccc}
\toprule
Method &Spider-dev & Spider-DK & Spider-Real & Spider-Syn \\ \midrule \midrule
FKD & 57.4 & 44.7 &\bf 52.8  &42.8 \\
 RKD &60.3	&50.5	&51.2	&44.6 \\ \hdashline
\texttt{KID} &\bf 63.7	&\bf 50.8	&52.2	&\bf 49.2 \\
\bottomrule
\end{tabular}
}
\caption{Performance (EX\%) on Spider-family benchmarks of QWen1.5-0.5B distilling from QWen1.5-32B.}
\label{tab:larger_teacher}
\end{table}

\paragraph{Does \texttt{KID} indeed alleviate the training-inference mismatch?} To verify it, we follow the prior work~\cite{gu2023knowledge} and use the ExAccErr~\cite{arora2022exposure} metric (lower score refers to less training-inference mismatch) to measure the training-inference mismatch. The results of QWen1.5-0.5B (distilling from QWen1.5-7B) on BIRD-dev (w/ EK) are listed in Table~\ref{tab:exaccerr}.
Obviously, comparing to the other methods, our \texttt{KID} achieves lower ExAccErr score, and there is a significant correlation between the ExAccErr score and the distillation performance, \textit{i.e.}, a lower mismatch leads to better performance. These results show the effectiveness of \texttt{KID}, and confirm our statement that alleviating the training-inference mismatch can enhance the distillation of text-to-SQL models.

\begin{table}[]
\centering
\scalebox{0.8}{
\begin{tabular}{lccccc}
\toprule
Metric &FKD & RKD &f-distill & GKD & \texttt{KID} \\ \midrule \midrule
ExAccErr ($\downarrow$) & 35.4 & 16.2 &11.3 &0.8  &5.3 \\
Performance &31.03	&31.81 &32.78	&34.62	&34.35 \\
\bottomrule
\end{tabular}
}
\caption{Results of Qwen1.5-0.5B on BIRD-dev (w/ EK) benchmark. QWen1.5-7B is used as the teacher.}
\label{tab:exaccerr}
\end{table}

%% file: Tables/main_results.tex
\begin{table*}[ht]
\centering
\scalebox{0.8}{
\begin{tabular}{lccccccccccccc}
\toprule
\multicolumn{1}{c}{\multirow{2}{*}{Method}} & \multirow{2}{*}{Latency} & \multicolumn{2}{c}{Spider-dev} & \multicolumn{2}{c}{BIRD-dev (EX\%)} & \multicolumn{2}{c}{Spider-DK} & \multicolumn{2}{c}{Spider-Real} & \multicolumn{2}{c}{Spider-Syn}  & \multicolumn{2}{c}{Score} \\ \cmidrule(lr){3-6} \cmidrule(lr){7-12} \cmidrule(lr){13-14}
 &  & EX\% & TS\% & w/o EK & w/ EK & EX\% & TS\% & EX\% & TS\% & EX\% & TS\%  & Avg. & $\Delta$ \\ \midrule \midrule
\multicolumn{14}{l}{\textit{Student: \textbf{QWen1.5-0.5B}}} \\
SFT & 1.0$\times$ & 57.8 & 56.4 & 16.36 & 30.51 & 44.8 & 46.5 & 50.6 & 47.6 & 44.2 & 43.7  & 43.85 & * \\  \midrule
\multicolumn{14}{l}{\textit{Teacher: \textbf{QWen1.5-1.8B}}} \\
Teacher & 1.5$\times$ & 67.3 & 66.3  & 21.71 & 34.22 & 54.6 & 52.3 & 62.0 & 60.8 & 52.7 & 52.6  & 52.45 & - \\ \hdashline
FKD & 2.1$\times$ & 57.3 & 56.5 & 16.82 & 28.68 & 43.7 & 41.7 & 50.2 & 48.0 & 43.7 & 43.3  & 42.99 & -0.86 \\
RKD & 2.0$\times$ & 62.7 & 61.5 & 16.10 & 31.81 & 50.8 & 49.2 & 51.2 & 49.6 & 48.7 & 48.3  & 46.99 & +3.14 \\
f-distill & 6.0$\times$ & 57.6 & 56.3 & 15.78 & 27.90 & 45.0 & 43.2 & 52.6 & 51.0 & 43.4 & 43.0  & 43.58 & -0.27 \\
ImitKD & 5.9$\times$ & 58.3 & 57.2 & 16.04 & 28.49  & 46.2 & 44.1 & 52.4 & 50.8 & 44.1 & 43.3 & 44.09 & +0.24 \\
GKD & 10.9$\times$ & 62.9 & 61.6 & 18.25 & 32.99 & \textbf{49.9} & \textbf{47.9} & 50.6 & 48.6 & \textbf{48.6} & \textbf{48.1}  & 46.94 & +3.09 \\
\rowcolor{gray!20} \textbf{\texttt{KID} (Ours)} & 2.0$\times$ & \textbf{63.7} & \textbf{63.1} & \textbf{18.38} & \textbf{33.12} & 47.6 & 45.4 & \textbf{53.0} & \textbf{51.4} & 47.5 & 47.0  & \textbf{47.02} & \textbf{+3.17} \\ \midrule
\multicolumn{14}{l}{\textit{Teacher: \textbf{QWen1.5-4B}}} \\
Teacher & 3.0$\times$ & 78.2 & 77.3 & 35.27 & 48.11 & 61.3 & 58.7 & 72.6 & 70.3 & 67.4 & 66.8  & 63.60 & - \\ \hdashline
FKD & 2.2$\times$ & 57.4 & 56.5 & 18.32 & 29.34 & 47.1 & 45.6 & 50.6 & 48.6 & 42.4 & 41.8  & 43.77 & -0.08 \\
RKD & 2.2$\times$ & 60.1 & 59.1 & 17.01 & 31.75 & 45.8 & 43.6 & 49.6 & 47.4 & 46.1 & 45.6  & 44.61 & +0.76 \\
f-distill & 6.3$\times$ & 58.6 & 57.3 & 17.67 & 31.55 & 45.8 & 43.6 & 50.8 & 49.2 & 44.4 & 43.8  & 44.27 & +0.42 \\
ImitKD & 6.3$\times$ & 59.5 & 59.4 & 19.04 & 30.31 & 48.6 & 46.9 & 49.2 & 46.9 & 45.0 & 44.5  & 44.94 & +1.09 \\
GKD & 12.7$\times$ & 63.8 & 62.4 & \textbf{20.21} & \textbf{36.11}  & \textbf{50.8} & \textbf{48.2} & \textbf{55.5} & \textbf{53.3} & 47.5 & 46.9 & 48.47 & +4.62 \\
\rowcolor{gray!20} \textbf{\texttt{KID} (Ours)} & 2.3$\times$ & \textbf{65.8} & \textbf{64.7} & 20.08 & 33.57  & 50.5 & 48.0 & 55.1 & \textbf{53.3} & \textbf{47.6} & \textbf{47.0} & \textbf{48.57} & \textbf{+4.72} \\ \midrule
\multicolumn{14}{l}{\textit{Teacher: \textbf{QWen1.5-7B}}} \\
Teacher & 3.3$\times$ & 81.6 & 80.6 & 39.44 & 52.02 & 67.7 & 64.9 & 76.6 & 74.2 & 70.1 & 69.5  & 67.67 & - \\ \hdashline
FKD & 2.4$\times$ & 57.3 & 56.4 & 17.14 & 31.03 & 46.4 & 44.9 & 50.6 & 49.0 & 41.0 & 40.5  & 43.43 & -0.42 \\
RKD & 2.3$\times$ & 61.5 & 60.2  & 16.10 & 31.81 & 48.4 & 46.5 & 51.0 & 49.2 & 46.7 & 46.0  & 45.74 & +1.89 \\
f-distill & 7.2$\times$ & 59.6 & 58.2 & 18.19 & 32.78 & 47.7 & 46.0 & 49.8 & 47.6 & 44.9 & 44.4  & 44.92 & +1.07 \\
ImitKD & 7.2$\times$ & 59.1 & 57.9 & 17.60 & 30.44  & 47.3 & 45.4 & 48.8 & 47.2 & 43.8 & 43.4 & 44.09 & +0.24 \\
GKD & 13.9$\times$ & \textbf{64.3} & \textbf{62.9} & 20.08 & \textbf{34.62} & \textbf{51.6} & \textbf{49.7} & \textbf{54.1} & \textbf{51.6} & 46.9 & 46.2  & \textbf{48.20} & \textbf{+4.35} \\
\rowcolor{gray!20} \textbf{\texttt{KID} (Ours)} & 2.3$\times$ & 64.0 & 62.6 & \textbf{20.40} & 34.35  & 50.7 & 48.5 & 52.4 & 50.8 & \textbf{47.7} & \textbf{47.3} & 47.88 & +4.03 \\
\bottomrule
\end{tabular}
}
\caption{\textbf{Evaluation of QWen-family models on several popular text-to-SQL benchmarks}. Notably, ``Latency'' means the average training latency relative to the SFT baseline. ``Spider-Real'' refers to the Spider-Realistic benchmark. ``Avg.'' denotes the average performance among all benchmarks and ``$\Delta$'' denotes the performance gains against the SFT baseline. Best performance in each group is emphasized in \textbf{bold}.}
\label{tab:main_qwen}
\end{table*}

%% file: Tables/main_results2.tex
\begin{table*}[ht]
\centering
\scalebox{0.8}{
\begin{tabular}{lccccccccccccc}
\toprule
\multicolumn{1}{c}{\multirow{2}{*}{Method}} & \multirow{2}{*}{Latency} & \multicolumn{2}{c}{Spider-dev} & \multicolumn{2}{c}{BIRD-dev (EX\%)} & \multicolumn{2}{c}{Spider-DK} & \multicolumn{2}{c}{Spider-Real} & \multicolumn{2}{c}{Spider-Syn}  & \multicolumn{2}{c}{Score} \\ \cmidrule(lr){3-6} \cmidrule(lr){7-12} \cmidrule(lr){13-14}
 &  & EX\% & TS\% & w/o EK & w/ EK & EX\% & TS\% & EX\% & TS\% & EX\% & TS\%  & Avg. & $\Delta$ \\ \midrule \midrule
 \multicolumn{14}{l}{\textit{Student: \textbf{CodeGen-350M}},  \textit{Teacher: \textbf{CodeGen-2B}}}. \\ \midrule
SFT & 1.0$\times$ & 53.1 & 51.8 & 9.90 & 26.01 & 37.4 & 36.1 & 38.4 & 36.0 & 35.4 & 34.9  & 35.90 & * \\ 
 % \multicolumn{14}{l}{\textit{Teacher Model: CodeGen-2B}} \\
Teacher & 3.7$\times$ & 72.3 & 71.3 & 26.47 & 35.66 & 57.9	&55.1	&63.2	&61.6	&55.4	&54.8  & 55.37 & - \\ \hdashline
% FKD & 2.0$\times$ & 53.2 & 52.2 & 40.7 & 39.1 & 40.6 & 38.4 & 36.4 & 36.0 & 10.30 & 24.58 & 37.15 & +1.25 \\
RKD & 2.1$\times$ & 55.1 & 54.4 & 10.50 & 27.18  & 43.6 & 40.0 & 43.1 & 40.7 & 37.6 & 36.8 & 38.90 & +3.00 \\
GKD & 14.1$\times$ & 56.6 & 54.9 &\bf 11.44 &\bf 27.57 & 43.7 & 40.4 & 45.5 & 43.1 & 40.1 & 39.3  & 40.26 & +4.36 \\
\rowcolor{gray!20} \textbf{\texttt{KID} (Ours)} & 2.4$\times$ &\bf 58.4 &\bf 56.8 & 10.52 &\bf 27.57 &\bf 46.4 &\bf 44.1 &\bf 47.6 &\bf 44.5 &\bf 41.1 &\bf 40.3  &\bf 41.73 & \bf +5.83 \\ 
% \hdashline
% \multicolumn{14}{l}{\textit{Teacher Model: CodeGen-6B}} \\ 
% Teacher & 4.8$\times$ & 73.9 & 72.6 & 60.9	&58.3	&66.3	&64.6	&60.3	&59.8 & 29.73 & 42.44 &58.89  & - \\
% FKD & 2.3$\times$ & 53.6 & 52.3 & 40.6 & 38.7 & 42.1 & 40.0 & 35.0 & 34.3 & 12.32 & 24.71 & 37.36 & +1.46 \\
% RKD & 2.4$\times$ & 55.9 & 54.9 & 42.4 & 39.4 & 42.1 & 40.0 & 38.1 & 37.4 & 13.82 & 26.21 & 39.02 & +3.12 \\
% GKD & 17.7$\times$ & 55.2 & 52.8 & 40.2 & 37.6 & 44.1 & 40.2 &\bf 39.9 &\bf 39.0 &\bf 14.43 & 26.50 & 38.99 & +3.09 \\
% \rowcolor{gray!20} \textbf{\texttt{KID} (Ours)} & 2.8$\times$ &\bf 58.3 &\bf 56.8 &\bf 46.2 &\bf 44.7 &\bf 45.7 &\bf 42.7 & 39.8 &\bf 39.0 & 14.41 &\bf 27.64 &\bf 41.53 & \bf +5.63 \\
\midrule \midrule
 \multicolumn{14}{l}{\textit{Student: \textbf{TinyLLaMA-1.1B}},  \textit{Teacher: \textbf{LLaMA2-7B}}}. \\ \midrule
SFT & 1.0$\times$ & 63.0 & 61.8 & 13.40 & 24.77 & 49.0 & 48.0 & 54.7 & 52.4 & 51.4 & 50.6  & 46.91 & * \\ 
% \multicolumn{14}{l}{\textit{Teacher Model: LLaMA2-7B}} \\ 
Teacher & 2.6$\times$ & 78.8 & 77.9 & 35.40 & 48.63 & 64.5	&61.1	&72.4	&70.1	&67.6	&66.4  &64.28  & - \\ \hdashline
% FKD & 1.4$\times$ & 63.2 & 61.8 & 46.4 & 45 & 54.7 & 52.8 & 50.0 & 48.8 & 14.15 & 31.03 & 46.79 & -0.12 \\
RKD & 1.4$\times$ & 66.0 & 64.6 & 15.45 & 31.75 & 48.4 & 46.9 & 55.7 & 54.1 & 52.9 & 52.2  & 48.80 & +1.89 \\
GKD & 8.3$\times$ & 64.8 & 63.2 & 16.62 &\bf 33.44 & 52.1 & 49.9 & 54.1 & 51.0 & 53.0 & 51.8  & 49.00 & +2.09 \\
\rowcolor{gray!20} \textbf{\texttt{KID} (Ours)} & 1.5$\times$ &\bf 68.1 &\bf 66.8 &\bf 18.97 & 32.53 &\bf 52.9 &\bf 51.8 &\bf 59.8 &\bf 57.7 &\bf 55.0 &\bf 54.5  & \bf51.81 & \bf +4.90 \\ 
% \hdashline
% \multicolumn{14}{l}{\textit{Teacher Model: LLaMA2-13B}} \\ 
% Teacher & 4.3$\times$ & 79.9 & 78.6 & 69.3	&65.2	&76.0	&74	&69.5	&68.6 & 39.83 & 51.89 &67.28  & - \\
% FKD & 1.9$\times$ & 62.1 & 60.8 & 47.9 & 47.1 & 53.1 & 51.4 & 48.5 & 47.8 & 14.28 & 31.94 & 46.49 & -0.42 \\
% RKD & 1.9$\times$ & 64.5 & 63.1 & 50.7 & 49.2 & 52.4 & 51.2 & 51.1 & 50.2 & 15.12 & 32.33 & 47.99 & +1.08 \\
% GKD & 8.8$\times$ &\bf 65.6 & 64.1 & 50.1 & 48.6 & 52.6 & 50.4 & 50.2 & 49.3 & 17.34 &\bf 33.12 & 48.14 & +1.23 \\
% \rowcolor{gray!20} \textbf{\texttt{KID} (Ours)} & 1.9$\times$ &\bf 65.6 &\bf 64.4 &\bf 51.0 &\bf 50.1 &\bf 56.5 &\bf 54.9 &\bf 54.2 &\bf 53.8 &\bf 18.84 & 32.46 &\bf 50.18 & \bf +3.27 \\
\bottomrule
\end{tabular}
}
\caption{\textbf{Evaluation of CodeGen and LLaMA models on several text-to-SQL benchmarks}. Due to the space constraints, we only present the contrastive results of most relevant KD counterparts, \textit{i.e.}, RKD and GKD.}
\label{tab:main_llama}
\end{table*}

% \begin{table}[]
% \centering
% \scalebox{0.8}{
% \begin{tabular}{lccccc}
% \toprule
% \multicolumn{1}{c}{\multirow{2}{*}{Method}} & \multicolumn{1}{c}{\multirow{2}{*}{Latency}} & \multicolumn{2}{c}{Spider-dev}& \multicolumn{2}{c}{BIRD-dev (EX\%)} \\ \cmidrule{3-6}
%  &  & EX\% & TS\% & w/o EK & w/ EK \\ \midrule \midrule
%  \multicolumn{6}{l}{\textit{Student: \textbf{CodeGen-350M}},  \textit{Teacher: \textbf{CodeGen-2B}}}. \\ \midrule
% SFT & 1.0$\times$ & 53.1 & 51.8 & 9.90 & 26.01 \\ 
%  % \multicolumn{14}{l}{\textit{Teacher Model: CodeGen-2B}} \\
% Teacher & 3.7$\times$ & 72.3 & 71.3 & 26.47 & 35.66 \\ \hdashline
% FKD & 2.0$\times$ & 53.2 & 52.2  & 10.30 & 24.58 \\
% RKD & 2.1$\times$ & 55.1 & 54.4 & 10.50 & 27.18  \\
% GKD & 14.1$\times$ & 56.6 & 54.9 &\bf 11.44 &\bf 27.57  \\
% \rowcolor{gray!20} \textbf{\texttt{KID} (Ours)} & 2.4$\times$ &\bf 58.4 &\bf 56.8 & 10.52 &\bf 27.57 \\ 
% \midrule \midrule
%  \multicolumn{6}{l}{\textit{Student: \textbf{TinyLLaMA-1.1B}},  \textit{Teacher: \textbf{LLaMA2-7B}}}. \\ \midrule
% SFT & 1.0$\times$ & 63.0 & 61.8 & 13.40 & 24.77 \\ 
% % \multicolumn{14}{l}{\textit{Teacher Model: LLaMA2-7B}} \\ 
% Teacher & 2.6$\times$ & 78.8 & 77.9 & 35.40 & 48.63  \\ \hdashline
% FKD & 1.4$\times$ & 63.2 & 61.8 & 14.15 & 31.03  \\
% RKD & 1.4$\times$ & 66.0 & 64.6 & 15.45 & 31.75 \\
% GKD & 8.3$\times$ & 64.8 & 63.2 & 16.62 &\bf 33.44 \\
% \rowcolor{gray!20} \textbf{\texttt{KID} (Ours)} & 1.5$\times$ &\bf 68.1 &\bf 66.8 &\bf 18.97 & 32.53 \\ 
% \bottomrule
% \end{tabular}
% }
% \end{table}

%% file: Section/5_related_work.tex
\section{Related Work}
\label{sec:related}

\paragraph{LLM-based Text-to-SQL.} Recently, autoregressive LLMs~\cite{openai2023gpt4,ouyang2022training,touvron2023llamav2,anil2023palm,zhao2023survey} have shown their superior performance by solving various NLP tasks in a generative manner. In the field of text-to-SQL, researchers are increasingly interested in leveraging the powerful capabilities of LLMs to create text-to-SQL systems, which can be classified into two groups: 1) prompt-based text-to-SQL and training-based text-to-SQL. The former involves designing some effective prompts to instruct the closed-source LLMs for better text-to-SQL parsing~\cite{pourreza2024din,sun2023sql,chen2023teaching,dong2023c3}. On the other hand, the training-based methods aim to improve the text-to-SQL performance of open-source LLMs by tuning them on the supervised input-output pairs~\cite{sun2023sql,zhang2024benchmarking}, or continuing pretraining the LLMs on the related database-related data~\cite{roziere2023code,li2024codes}. While achieving remarkable performance, the above methods usually suffer from unbearable inference latency~\cite{zhong2024revisiting,leviathan2023fast}, hindering the applications in real-world scenarios.
% Hence, it is crucial and green to explore the model compression techniques for accelerating the inference, while not losing much performance~\cite{schwartz2020green}.

\paragraph{Knowledge Distillation for Autoregressive LLMs.} KD, as a common approach for compressing LLMs, has attracted great attention recently~\cite{gu2023knowledge,agarwal2023gkd,zhong2024revisiting,rao2024exploring,xu2024survey}. In the context of text-to-SQL, \citet{sun2023exploratory} is first to apply the KD for distilling the text-to-SQL models, but they mainly focus on the encoder-only~\cite{devlin2019bert} and sequence-to-sequence models~\cite{raffel2020exploring}. It is still under-explored whether these methods work well for distilling autoregressive text-to-SQL LLMs. In this paper, we conduct a series of preliminary experiments to explore it and reveal that training-inference mismatch is one of the main factors hindering the KD performance in autoregressive LLMs. Hence, we propose an effective and efficient KD method to alleviate the training-inference mismatch. Notably, our motivation is similar to the schedule sampling~\cite{bengio2015scheduled}, but there are significant differences between the two. We depart from the prior schedule sampling and ours as follows: 1) \textit{Different approaches}: schedule sampling focuses on RNN models involving serial training, whereas ours targets Transformer models requiring parallel training. 2) \textit{Different application scenarios}: schedule sampling was applied to small RNN model training, but our method is applied in the distillation scenario of LLMs, especially for the text-to-SQL. 
% 3) \textit{Different focal points}: schedule sampling mainly aims at improving training performance, while our work focuses on achieving a better performance-efficiency trade-off, with an emphasis on enhancing training efficiency. 
% Hence, we attempt to explore it and propose a more efficient KD method that is more suitable for text-to-SQL LLMs. Specifically, we reveal that the training-inference mismatch is one of the main factors for hindering the KD performance. Thus, a 

% To the best of our knowledge, we are one of the rare works that focus on efficient LLM-based text-to-SQL systems, and we hope our work can promote more related research in this field.

% we first conduct the comprehensive evaluation of the representative KD methods on autoregressive LLMs and find that these methods fail to achieve well trade-off between performance and efficiency. To this end, we then propose our novel \texttt{KID} method, which can achieve remarkable performance without introducing much training latency.

%% file: Section/6_conclusion.tex
\section{Conclusion}
In this paper, we reveal and address the limitations of current KD methods in compressing the autoregressive text-to-SQL LLMs. Based on a series of preliminary analyses, we find that these methods fall short in balancing performance and training efficiency. 
% Specifically, although those based on model-generated data can alleviate the training-inference gap and perform well, they usually suffer from training inefficiency. 
To this end, we propose a novel efficient KD algorithm (\texttt{KID}), which utilizes a simple-yet-effective strategy to simulate the inference scenarios during training, with only a one-pass forward process. By doing so, \texttt{KID} can mitigate the training-inference mismatch in an efficient manner, and achieve a better trade-off between performance and efficiency. Experiments show that our approach consistently and significantly improves distillation performance across all model architectures, and reduces the training latency by a large margin.

\section*{Limitations}
Our work has several potential limitations.
First, given the limited computational budget, we only validate our \texttt{KID} on up to 7B LLMs in the main experiments. It will be more convincing if scaling up to super-large model size, \textit{e.g.}, 70B.
% and applying \texttt{KID} to more cutting-edge model architectures. 
Secondly, in our \texttt{KID}, we leverage an auxiliary MLE loss to ensure the stable training. In our preliminary experiments, we found that the MLE loss plays an import role in \texttt{KID}. However, the better combination of the distillation loss and MLE loss is still under-explored, which is in our future work. Lastly, besides the distillation for text-to-SQL, we believe that our method has the great potential to expand to more scenarios. 
% \textit{e.g.}, distilling the general-purpose abilities of LLMs.

\section*{Ethics Statements} 
We take ethical considerations very seriously and strictly adhere to the ACL Ethics Policy. This paper proposes an efficient knowledge distillation algorithm for text-to-SQL LLMs. It aims to compress the existing larger LLMs into smaller ones, instead of encouraging them to learn privacy knowledge that may cause the ethical problem. Moreover, all training and evaluation datasets used in this paper are publicly available and have been widely adopted by researchers. Thus, we believe that this research will not pose ethical issues.

% \paragraph{Reproducibility.}
% In this paper, we discuss the detailed experimental setup and provide enough information to re-product our results, such as statistic descriptions and training hyper-parameters. More importantly, we will publicly release our code in \url{https://github.com/WHU-ZQH/KID} to help reproduce the experimental results of this paper.

\section*{Acknowledgements}
We are grateful to the anonymous reviewers and the area chair for their insightful comments and suggestions.
This work was supported in part by the National Natural Science Foundation of China under Grant 623B2076, U23B2048, 62076186 and 62225113, in part by the National Key Research and Development Program of China under Grant 2023YFC2705700, in part by the Innovative Research Group Project of Hubei Province under Grant 2024AFA017, and in part by the National Research Foundation, Singapore, and the CyberSG R\&D Programme Office (``CRPO''),  under the National Cybersecurity R\&D Programme (``NCRP''), RIE2025 NCRP Funding Initiative (Award CRPO-GC1-NTU-002). The numerical calculations in this paper have been done on the supercomputing system in the Supercomputing Center of Wuhan University.

%% file: Section/7_appendix.tex
\appendix
\section{Appendix}
\label{sec:appendix}

\subsection{Details of Tasks and Datasets}
\label{appendix_data}
In this work, we conduct extensive experiments on several text-to-SQL benchmarks. Here, we introduce the descriptions of these datasets in detail. Firstly, we present the statistics of all used datasets in Table~\ref{tab:dataset}. Then, each task is described as:

\textbf{Spider.} Spider~\cite{yu2018spider} is a widely-used English text-to-SQL benchmark, comprising 8,659 training samples and 1,034 development samples. The training set encompasses 7,000 manually annotated samples and 1,659 samples sourced from six previous text-to-SQL benchmarks. There are 200 databases covering 138 diverse domains in Spider. Due to the submission constraints of the Spider leaderboard, we follow~\citet{li2024codes} and do not evaluate our models on its test set, but alternatively on the publicly available development set.

\textbf{BIRD.} BIRD~\cite{li2024can} is a more challenging text-to-SQL benchmark that examines the impact of extensive database contents on text-to-SQL parsing. BIRD contains over 12,751 unique question-SQL pairs and 95 big databases with a total size of 33.4 GB. Each database contains around 549K rows on average. 

\textbf{Spider-DK.} Spider-DK~\cite{gan2021exploring} is a variant derived from the original Spider dataset. It modifies some samples of Spider by adding domain knowledge that reflects real-world question paraphrases. 

\textbf{Spider-Realistic.} Spider-Realistic~\cite{deng2021structure} is also a variant of Spider dataset. It modifies the NL questions in the complex subset of Spider to remove or paraphrase explicit mentions of column names, while keeping the SQL queries unchanged.

\textbf{Spider-Syn.} Spider-Syn~\cite{gan2021towards} is a human-curated dataset based on the Spider. NL questions in Spider-Syn are modified from Spider, by replacing their schema-related words with manually selected synonyms that reflect real-world question para-phrases.

\begin{table}[t]
\centering
\setlength{\tabcolsep}{10pt}
\scalebox{0.9}{
\begin{tabular}{lcc}
\toprule
\bf Benchmark &\bf \#Training & \bf \#Development \\ \midrule
Spider & 8,659 & 1,034 \\
BIRD & 9,428 & 1,534 \\
Spider-DK & - & 535 \\
Spider-Realistic & - & 508 \\
Spider-Syn & - & 1,034 \\
\bottomrule
\end{tabular}
}
\caption{\textbf{Statistic of all used text-to-SQL benchmarks}. Notably, ``Spider-DK'', ``Spider-Realistic'' and ``Spider-Syn'' are variants of the development of Spider.}
\label{tab:dataset}
\end{table}

\begin{table}[]
\centering
\scalebox{0.82}{
\begin{tabular}{lccc}
\toprule
\bf Setting &\bf  QWen1.5 & \bf CodeGen & \bf LLaMA2 \\ \midrule
Learning Rate & 2e-4 & 2e-4 & 2e-4 \\
Epoch & 8 & 8 & 4 \\
Batch Size & 16 & 16 & 16 \\
Max Input Length & 1024 & 1024 & 2048 \\
Max Output Length & 128 & 128 & 256 \\
LoRA\_Rank & 64 & 8 & 64 \\
LoRA\_Alpha & 32 & 32 & 32 \\
\bottomrule
\end{tabular}
}
\caption{\textbf{Details of training hyper-parameters for different LLMs}. For each model, we use the same settings among all benchmarks.}
\label{tab:training_details}
\end{table}

\begin{figure}[t]
\centering
    \includegraphics[width=0.48\textwidth]{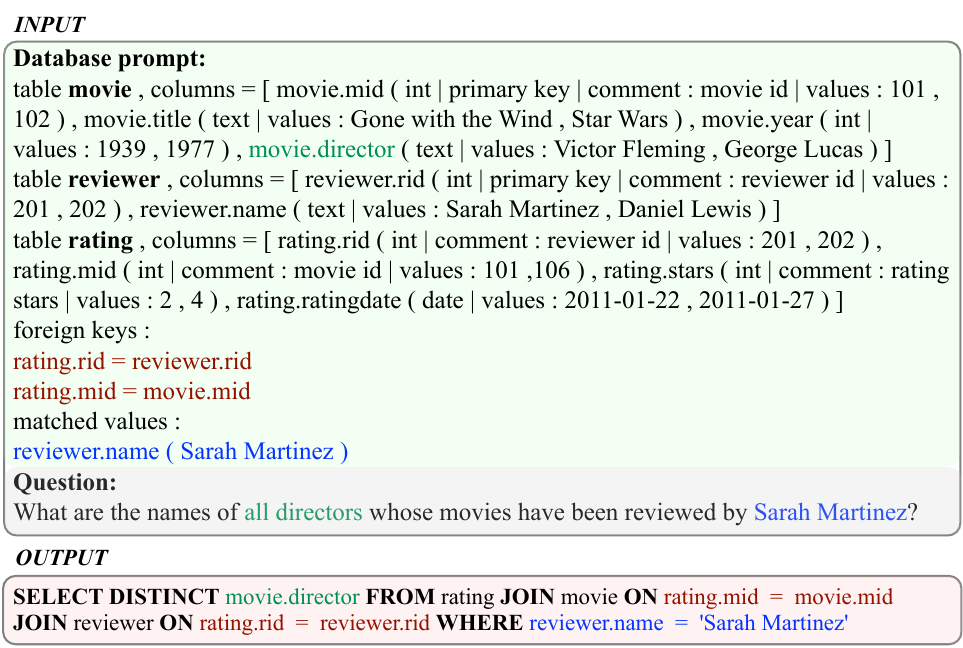}
    \caption{\textbf{A text-to-SQL sample in Spider's training set}. We follow~\citet{li2024codes} to construct the database prompts. Note that this illustration is from the original paper~\cite{li2024codes}.}
    \label{figure:prompt}
\end{figure}

\subsection{Training Hyper-parameters.}
\label{appendix_training}
We train each model with a batch size of 16 and a peak learning rate of 2e-4. The training epochs are selected from \{4, 8\} for different models. We follow~\citet{li2024codes} to construct the database prompt (an example of an input-output pair is illustrated in Figure~\ref{figure:prompt}) and set the max length of input and output depending on different models. Due to the limited computational resources, we train all models with a popular parameter-efficient fine-tuning method, \textit{i.e.}, LoRA. Specifically, the alpha of LoRA is set as 32 and the rank of LoRA is set as 64 or 8. We present the training hyper-parameters in Table~\ref{tab:training_details}. All experiments are conducted on 8 NVIDIA H800 (80GB) GPUs.

\subsection{Details of divergence functions for KD}
\label{appendix:divergence_fn}
Here, we introduce the commonly-used divergence functions for KD. Let the probability distribution of teacher and student be $p$ and $q^\theta$, respectively. For the training set $\mathcal{G}$, the divergence functions can be formulated as:

\paragraph{Kullback-Leibler~(KL) divergence}
\begin{equation}
    \mathcal{F}_{KL}(p\|q^\theta) = \sum_{(x,y) \in \mathcal{G}} p(y|x) \log \frac{p(y|x)}{q^\theta(y|x)}.
\end{equation}

Note that the KL divergence is not symmetric, \textit{i.e.}, $\mathcal{F}_{KL}(p\|q^\theta) \neq \mathcal{F}_{KL}(q^\theta\|p)$. More specifically, the $\mathcal{F}_{KL}(p\|q^\theta)$ refers to the forward KL, while $\mathcal{F}_{KL}(q^\theta\|p)$ refers to the reverse KL.

\paragraph{Jensen–Shannon~(JS) divergence}
\begin{equation}
    \mathcal{F}_{JS} (p\|q^\theta) = \frac{1}{2}(\mathcal{F}_{KL}(p\|M) +\mathcal{F}_{KL}(q^\theta\|M)),
\end{equation}
where $M = \frac{1}{2}(p+q^\theta)$. 

\paragraph{Total variation distance~(TVD)}
\begin{equation}
    \mathcal{F}_{TVD}(p\|q^\theta) = \sum_{(x,y) \in \mathcal{G}}{|\frac{p(y|x) -q^\theta(y|x)}{2}|}.
\end{equation}

%% file: acl.bbl
\begin{thebibliography}{50}
\providecommand{\natexlab}[1]{#1}

\bibitem[{Agarwal et~al.(2024)Agarwal, Vieillard, Stanczyk, Ramos, Geist, and Bachem}]{agarwal2023gkd}
Rishabh Agarwal, Nino Vieillard, Piotr Stanczyk, Sabela Ramos, Matthieu Geist, and Olivier Bachem. 2024.
\newblock \href {https://openreview.net/forum?id=3zKtaqxLhW} {On-policy distillaiton of language models: Learning from self-generated mistakes}.
\newblock In \emph{ICLR}.

\bibitem[{Anil et~al.(2023)Anil, Dai, Firat, Johnson, Lepikhin, Passos, Shakeri, Taropa, Bailey, Chen et~al.}]{anil2023palm}
Rohan Anil, Andrew~M Dai, Orhan Firat, Melvin Johnson, Dmitry Lepikhin, Alexandre Passos, Siamak Shakeri, Emanuel Taropa, Paige Bailey, Zhifeng Chen, et~al. 2023.
\newblock \href {https://arxiv.org/pdf/2305.10403.pdf} {Palm 2 technical report}.
\newblock \emph{arXiv preprint}.

\bibitem[{Arora et~al.(2022)Arora, El~Asri, Bahuleyan, and Cheung}]{arora2022exposure}
Kushal Arora, Layla El~Asri, Hareesh Bahuleyan, and Jackie Chi~Kit Cheung. 2022.
\newblock \href {https://aclanthology.org/2022.findings-acl.58/} {Why exposure bias matters: An imitation learning perspective of error accumulation in language generation}.
\newblock In \emph{Findings of ACL 2022}.

\bibitem[{Bai et~al.(2023)Bai, Bai, Chu, Cui, Dang, Deng, Fan, Ge, Han, Huang et~al.}]{bai2023qwen}
Jinze Bai, Shuai Bai, Yunfei Chu, Zeyu Cui, Kai Dang, Xiaodong Deng, Yang Fan, Wenbin Ge, Yu~Han, Fei Huang, et~al. 2023.
\newblock \href {https://arxiv.org/pdf/2309.16609.pdf?trk=public_post_comment-text} {Qwen technical report}.
\newblock \emph{arXiv preprint}.

\bibitem[{Bengio et~al.(2015)Bengio, Vinyals, Jaitly, and Shazeer}]{bengio2015scheduled}
Samy Bengio, Oriol Vinyals, Navdeep Jaitly, and Noam Shazeer. 2015.
\newblock \href {https://proceedings.neurips.cc/paper/2015/file/e995f98d56967d946471af29d7bf99f1-Paper.pdf} {Scheduled sampling for sequence prediction with recurrent neural networks}.
\newblock In \emph{NeurIPS}.

\bibitem[{Chen et~al.(2024)Chen, Lin, Schaerli, and Zhou}]{chen2023teaching}
Xinyun Chen, Maxwell Lin, Nathanael Schaerli, and Denny Zhou. 2024.
\newblock \href {https://openreview.net/forum?id=KuPixIqPiq} {Teaching large language models to self-debug}.
\newblock In \emph{ICLR}.

\bibitem[{Deng et~al.(2021)Deng, Hassan, Meek, Polozov, Sun, and Richardson}]{deng2021structure}
Xiang Deng, Ahmed Hassan, Christopher Meek, Oleksandr Polozov, Huan Sun, and Matthew Richardson. 2021.
\newblock \href {https://aclanthology.org/2021.naacl-main.105/} {Structure-grounded pretraining for text-to-sql}.
\newblock In \emph{NAACL}.

\bibitem[{Devlin et~al.(2019)Devlin, Chang, Lee, and Toutanova}]{devlin2019bert}
Jacob Devlin, Ming-Wei Chang, Kenton Lee, and Kristina Toutanova. 2019.
\newblock Bert: Pre-training of deep bidirectional transformers for language understanding.
\newblock In \emph{NAACL}.

\bibitem[{Dong et~al.(2023)Dong, Zhang, Ge, Mao, Gao, Lin, Lou et~al.}]{dong2023c3}
Xuemei Dong, Chao Zhang, Yuhang Ge, Yuren Mao, Yunjun Gao, Jinshu Lin, Dongfang Lou, et~al. 2023.
\newblock \href {https://arxiv.org/pdf/2307.07306} {C3: Zero-shot text-to-sql with chatgpt}.
\newblock \emph{arXiv preprint}.

\bibitem[{Fuglede and Topsoe(2004)}]{fuglede2004jensen}
Bent Fuglede and Flemming Topsoe. 2004.
\newblock \href {https://www.researchgate.net/profile/Flemming-Topsoe/publication/4109004_Jensen-Shannon_Divergence_and_Hilbert_Space_Embedding/links/0c9605249611e4c842000000/Jensen-Shannon-Divergence-and-Hilbert-Space-Embedding.pdf} {Jensen-shannon divergence and hilbert space embedding}.
\newblock In \emph{International symposium onInformation theory, 2004. ISIT 2004. Proceedings.}

\bibitem[{Gan et~al.(2021{\natexlab{a}})Gan, Chen, Huang, Purver, Woodward, Xie, and Huang}]{gan2021towards}
Yujian Gan, Xinyun Chen, Qiuping Huang, Matthew Purver, John~R Woodward, Jinxia Xie, and Pengsheng Huang. 2021{\natexlab{a}}.
\newblock \href {https://aclanthology.org/2021.acl-long.195/} {Towards robustness of text-to-sql models against synonym substitution}.
\newblock In \emph{ACL}.

\bibitem[{Gan et~al.(2021{\natexlab{b}})Gan, Chen, and Purver}]{gan2021exploring}
Yujian Gan, Xinyun Chen, and Matthew Purver. 2021{\natexlab{b}}.
\newblock \href {https://aclanthology.org/2021.emnlp-main.702/} {Exploring underexplored limitations of cross-domain text-to-sql generalization}.
\newblock In \emph{EMNLP}.

\bibitem[{Gu et~al.(2023)Gu, Dong, Wei, and Huang}]{gu2023knowledge}
Yuxian Gu, Li~Dong, Furu Wei, and Minlie Huang. 2023.
\newblock \href {https://arxiv.org/pdf/2306.08543} {Knowledge distillation of large language models}.
\newblock \emph{arXiv preprint}.

\bibitem[{Hinton et~al.(2015)Hinton, Vinyals, and Dean}]{hinton2015distilling}
Geoffrey Hinton, Oriol Vinyals, and Jeff Dean. 2015.
\newblock \href {https://arxiv.org/pdf/1503.02531} {Distilling the knowledge in a neural network}.
\newblock \emph{arXiv preprint}.

\bibitem[{Hu et~al.(2021)Hu, Wallis, Allen-Zhu, Li, Wang, Wang, Chen et~al.}]{hu2021lora}
Edward~J Hu, Phillip Wallis, Zeyuan Allen-Zhu, Yuanzhi Li, Shean Wang, Lu~Wang, Weizhu Chen, et~al. 2021.
\newblock \href {https://arxiv.org/abs/2106.09685} {Lora: Low-rank adaptation of large language models}.
\newblock In \emph{ICLR}.

\bibitem[{Katsogiannis-Meimarakis and Koutrika(2023)}]{katsogiannis2023survey}
George Katsogiannis-Meimarakis and Georgia Koutrika. 2023.
\newblock \href {https://link.springer.com/content/pdf/10.1007/s00778-022-00776-8.pdf} {A survey on deep learning approaches for text-to-sql}.
\newblock \emph{The VLDB Journal}.

\bibitem[{Kim and Rush(2016)}]{kim2016sequence}
Yoon Kim and Alexander~M Rush. 2016.
\newblock \href {https://aclanthology.org/D16-1139.pdf} {Sequence-level knowledge distillation}.
\newblock In \emph{EMNLP}.

\bibitem[{Leviathan et~al.(2023)Leviathan, Kalman, and Matias}]{leviathan2023fast}
Yaniv Leviathan, Matan Kalman, and Yossi Matias. 2023.
\newblock \href {https://proceedings.mlr.press/v202/leviathan23a/leviathan23a.pdf} {Fast inference from transformers via speculative decoding}.
\newblock In \emph{ICML}.

\bibitem[{Li et~al.(2023)Li, Zhang, Li, and Chen}]{li2023resdsql}
Haoyang Li, Jing Zhang, Cuiping Li, and Hong Chen. 2023.
\newblock \href {https://ojs.aaai.org/index.php/AAAI/article/download/26535/26307} {Resdsql: Decoupling schema linking and skeleton parsing for text-to-sql}.
\newblock In \emph{AAAI}.

\bibitem[{Li et~al.(2024{\natexlab{a}})Li, Zhang, Liu, Fan, Zhang, Zhu, Wei, Pan, Li, and Chen}]{li2024codes}
Haoyang Li, Jing Zhang, Hanbing Liu, Ju~Fan, Xiaokang Zhang, Jun Zhu, Renjie Wei, Hongyan Pan, Cuiping Li, and Hong Chen. 2024{\natexlab{a}}.
\newblock \href {https://arxiv.org/pdf/2402.16347} {Codes: Towards building open-source language models for text-to-sql}.
\newblock \emph{Proceedings of the ACM on Management of Data}.

\bibitem[{Li et~al.(2024{\natexlab{b}})Li, Hui, Qu, Yang, Li, Li, Wang, Qin, Geng, Huo et~al.}]{li2024can}
Jinyang Li, Binyuan Hui, Ge~Qu, Jiaxi Yang, Binhua Li, Bowen Li, Bailin Wang, Bowen Qin, Ruiying Geng, Nan Huo, et~al. 2024{\natexlab{b}}.
\newblock \href {https://proceedings.neurips.cc/paper_files/paper/2023/file/83fc8fab1710363050bbd1d4b8cc0021-Paper-Datasets_and_Benchmarks.pdf} {Can llm already serve as a database interface? a big bench for large-scale database grounded text-to-sqls}.
\newblock In \emph{NeurIPS}.

\bibitem[{Lin et~al.(2020)Lin, Wohlwend, Chen, and Lei}]{lin2020autoregressive}
Alexander Lin, Jeremy Wohlwend, Howard Chen, and Tao Lei. 2020.
\newblock \href {https://aclanthology.org/2020.emnlp-main.494.pdf} {Autoregressive knowledge distillation through imitation learning}.
\newblock In \emph{EMNLP}.

\bibitem[{Liu et~al.(2023)Liu, Hu, Bailis, Stoica, Deng, Cheung, and Zhang}]{liu2023online}
Xiaoxuan Liu, Lanxiang Hu, Peter Bailis, Ion Stoica, Zhijie Deng, Alvin Cheung, and Hao Zhang. 2023.
\newblock \href {https://arxiv.org/pdf/2310.07177} {Online speculative decoding}.
\newblock In \emph{ICLR}.

\bibitem[{Malinin and Gales(2019)}]{malinin2019reverse}
Andrey Malinin and Mark Gales. 2019.
\newblock \href {https://proceedings.neurips.cc/paper/2019/file/7dd2ae7db7d18ee7c9425e38df1af5e2-Paper.pdf} {Reverse kl-divergence training of prior networks: Improved uncertainty and adversarial robustness}.
\newblock \emph{NeurIPS}.

\bibitem[{Nijkamp et~al.(2022)Nijkamp, Pang, Hayashi, Tu, Wang, Zhou, Savarese, and Xiong}]{nijkamp2022codegen}
Erik Nijkamp, Bo~Pang, Hiroaki Hayashi, Lifu Tu, Huan Wang, Yingbo Zhou, Silvio Savarese, and Caiming Xiong. 2022.
\newblock \href {https://arxiv.org/abs/2203.13474} {Codegen: An open large language model for code with multi-turn program synthesis}.
\newblock In \emph{ICLR}.

\bibitem[{OpenAI(2023)}]{openai2023gpt4}
OpenAI. 2023.
\newblock \href {https://arxiv.org/abs/2303.08774} {Gpt-4 technical report}.
\newblock \emph{Preprint}, arXiv preprint:2303.08774.

\bibitem[{Ouyang et~al.(2022)Ouyang, Wu, Jiang, Almeida, Wainwright, Mishkin, Zhang, Agarwal, Slama, Ray et~al.}]{ouyang2022training}
Long Ouyang, Jeffrey Wu, Xu~Jiang, Diogo Almeida, Carroll Wainwright, Pamela Mishkin, Chong Zhang, Sandhini Agarwal, Katarina Slama, Alex Ray, et~al. 2022.
\newblock \href {https://proceedings.neurips.cc/paper_files/paper/2022/file/b1efde53be364a73914f58805a001731-Paper-Conference.pdf} {Training language models to follow instructions with human feedback}.
\newblock In \emph{NeurIPS}.

\bibitem[{Pang and He(2020)}]{pang2020text}
Richard~Yuanzhe Pang and He~He. 2020.
\newblock Text generation by learning from demonstrations.
\newblock In \emph{ICLR}.

\bibitem[{Pourreza and Rafiei(2024)}]{pourreza2024din}
Mohammadreza Pourreza and Davood Rafiei. 2024.
\newblock \href {https://proceedings.neurips.cc/paper_files/paper/2023/file/72223cc66f63ca1aa59edaec1b3670e6-Paper-Conference.pdf} {Din-sql: Decomposed in-context learning of text-to-sql with self-correction}.
\newblock In \emph{NeurIPS}.

\bibitem[{Raffel et~al.(2020)Raffel, Shazeer, Roberts, Lee, Narang, Matena, Zhou, Li, and Liu}]{raffel2020exploring}
Colin Raffel, Noam Shazeer, Adam Roberts, Katherine Lee, Sharan Narang, Michael Matena, Yanqi Zhou, Wei Li, and Peter~J Liu. 2020.
\newblock \href {https://www.jmlr.org/papers/volume21/20-074/20-074.pdf} {Exploring the limits of transfer learning with a unified text-to-text transformer}.
\newblock \emph{JMLR}.

\bibitem[{Rao et~al.(2024)Rao, Liu, Lin, Ding, Li, and Tao}]{rao2024exploring}
Jun Rao, Xuebo Liu, Zepeng Lin, Liang Ding, Jing Li, and Dacheng Tao. 2024.
\newblock \href {https://arxiv.org/abs/2409.12512} {Exploring and enhancing the transfer of distribution in knowledge distillation for autoregressive language models}.
\newblock \emph{arXiv preprint}.

\bibitem[{Roziere et~al.(2023)Roziere, Gehring, Gloeckle, Sootla, Gat, Tan, Adi, Liu, Remez, Rapin et~al.}]{roziere2023code}
Baptiste Roziere, Jonas Gehring, Fabian Gloeckle, Sten Sootla, Itai Gat, Xiaoqing~Ellen Tan, Yossi Adi, Jingyu Liu, Tal Remez, J{\'e}r{\'e}my Rapin, et~al. 2023.
\newblock \href {https://arxiv.org/abs/2308.12950} {Code llama: Open foundation models for code}.
\newblock \emph{arXiv preprint}.

\bibitem[{Schwartz et~al.(2020)Schwartz, Dodge, Smith, and Etzioni}]{schwartz2020green}
Roy Schwartz, Jesse Dodge, Noah~A Smith, and Oren Etzioni. 2020.
\newblock \href {https://dl.acm.org/doi/pdf/10.1145/3381831} {Green ai}.
\newblock \emph{Communications of the ACM}.

\bibitem[{Sun et~al.(2023{\natexlab{a}})Sun, Arik, Nakhost, Dai, Sinha, Yin, and Pfister}]{sun2023sql}
Ruoxi Sun, Sercan~O Arik, Hootan Nakhost, Hanjun Dai, Rajarishi Sinha, Pengcheng Yin, and Tomas Pfister. 2023{\natexlab{a}}.
\newblock \href {https://arxiv.org/pdf/2306.00739.pdf} {Sql-palm: Improved large language modeladaptation for text-to-sql}.
\newblock \emph{arXiv preprint}.

\bibitem[{Sun et~al.(2023{\natexlab{b}})Sun, Gao, Zhang, Su, Chen, Lin, and Sun}]{sun2023exploratory}
Shuo Sun, Yuze Gao, Yuchen Zhang, Jian Su, Bin Chen, Yingzhan Lin, and Shuqi Sun. 2023{\natexlab{b}}.
\newblock \href {https://aclanthology.org/2023.findings-acl.740.pdf} {An exploratory study on model compression for text-to-sql}.
\newblock In \emph{Findings of ACL}.

\bibitem[{Touvron et~al.(2023)Touvron, Martin, Stone, Albert, Almahairi, Babaei, Bashlykov, Batra, Bhargava, Bhosale et~al.}]{touvron2023llamav2}
Hugo Touvron, Louis Martin, Kevin Stone, Peter Albert, Amjad Almahairi, Yasmine Babaei, Nikolay Bashlykov, Soumya Batra, Prajjwal Bhargava, Shruti Bhosale, et~al. 2023.
\newblock \href {https://arxiv.org/pdf/2307.09288.pdf%C3%82%C2%A0} {Llama 2: Open foundation and fine-tuned chat models}.
\newblock \emph{arXiv preprint}.

\bibitem[{Van~Erven and Harremos(2014)}]{van2014renyi}
Tim Van~Erven and Peter Harremos. 2014.
\newblock \href {https://arxiv.org/pdf/1206.2459} {R{\'e}nyi divergence and kullback-leibler divergence}.
\newblock \emph{IEEE Transactions on Information Theory}.

\bibitem[{Verd{\'u}(2014)}]{verdu2014total}
Sergio Verd{\'u}. 2014.
\newblock \href {http://ita.ucsd.edu/workshop/14/files/paper/paper_374.pdf} {Total variation distance and the distribution of relative information}.
\newblock In \emph{2014 Information Theory and Applications Workshop (ITA)}.

\bibitem[{Wen et~al.(2023)Wen, Li, Du, and Mou}]{wen2023f}
Yuqiao Wen, Zichao Li, Wenyu Du, and Lili Mou. 2023.
\newblock \href {https://aclanthology.org/2023.acl-long.605.pdf} {f-divergence minimization for sequence-level knowledge distillation}.
\newblock In \emph{ACL}.

\bibitem[{Wu et~al.(2024)Wu, Tao, Wang, Zhao, and Wong}]{wu2024rethinking}
Taiqiang Wu, Chaofan Tao, Jiahao Wang, Zhe Zhao, and Ngai Wong. 2024.
\newblock \href {https://arxiv.org/pdf/2404.02657} {Rethinking kullback-leibler divergence in knowledge distillation for large language models}.
\newblock \emph{arXiv preprint}.

\bibitem[{Xu et~al.(2024)Xu, Li, Tao, Shen, Cheng, Li, Xu, Tao, and Zhou}]{xu2024survey}
Xiaohan Xu, Ming Li, Chongyang Tao, Tao Shen, Reynold Cheng, Jinyang Li, Can Xu, Dacheng Tao, and Tianyi Zhou. 2024.
\newblock \href {https://arxiv.org/pdf/2402.13116} {A survey on knowledge distillation of large language models}.
\newblock \emph{arXiv preprint}.

\bibitem[{Yu et~al.(2018)Yu, Zhang, Yang, Yasunaga, Wang, Li, Ma, Li, Yao, Roman et~al.}]{yu2018spider}
Tao Yu, Rui Zhang, Kai Yang, Michihiro Yasunaga, Dongxu Wang, Zifan Li, James Ma, Irene Li, Qingning Yao, Shanelle Roman, et~al. 2018.
\newblock \href {https://aclanthology.org/D18-1425/} {Spider: A large-scale human-labeled dataset for complex and cross-domain semantic parsing and text-to-sql task}.
\newblock In \emph{EMNLP}.

\bibitem[{Zhang et~al.(2024{\natexlab{a}})Zhang, Ye, Du, Hu, Li, Yang, Liu, Zhao, Li, and Mao}]{zhang2024benchmarking}
Bin Zhang, Yuxiao Ye, Guoqing Du, Xiaoru Hu, Zhishuai Li, Sun Yang, Chi~Harold Liu, Rui Zhao, Ziyue Li, and Hangyu Mao. 2024{\natexlab{a}}.
\newblock \href {https://arxiv.org/pdf/2403.02951} {Benchmarking the text-to-sql capability of large language models: A comprehensive evaluation}.
\newblock \emph{arXiv preprint}.

\bibitem[{Zhang et~al.(2024{\natexlab{b}})Zhang, Zeng, Wang, and Lu}]{zhang2024tinyllama}
Peiyuan Zhang, Guangtao Zeng, Tianduo Wang, and Wei Lu. 2024{\natexlab{b}}.
\newblock \href {https://arxiv.org/pdf/2401.02385} {Tinyllama: An open-source small language model}.
\newblock \emph{arXiv preprint}.

\bibitem[{Zhao et~al.(2023)Zhao, Zhou, Li, Tang, Wang, Hou, Min, Zhang, Zhang, Dong et~al.}]{zhao2023survey}
Wayne~Xin Zhao, Kun Zhou, Junyi Li, Tianyi Tang, Xiaolei Wang, Yupeng Hou, Yingqian Min, Beichen Zhang, Junjie Zhang, Zican Dong, et~al. 2023.
\newblock \href {https://arxiv.org/pdf/2303.18223.pdf,} {A survey of large language models}.
\newblock \emph{arXiv preprint}.

\bibitem[{Zhong et~al.(2023)Zhong, Ding, Liu, Du, and Tao}]{zhong2023self}
Qihuang Zhong, Liang Ding, Juhua Liu, Bo~Du, and Dacheng Tao. 2023.
\newblock \href {https://aclanthology.org/2023.findings-acl.254/} {Self-evolution learning for discriminative language model pretraining}.
\newblock In \emph{Findings of ACL}.

\bibitem[{Zhong et~al.(2024)Zhong, Ding, Shen, Liu, Du, and Tao}]{zhong2024revisiting}
Qihuang Zhong, Liang Ding, Li~Shen, Juhua Liu, Bo~Du, and Dacheng Tao. 2024.
\newblock \href {https://arxiv.org/pdf/2402.11890} {Revisiting knowledge distillation for autoregressive language models}.
\newblock In \emph{ACL}.

\bibitem[{Zhong et~al.(2020)Zhong, Yu, and Klein}]{zhong2020semantic}
Ruiqi Zhong, Tao Yu, and Dan Klein. 2020.
\newblock \href {https://aclanthology.org/2020.emnlp-main.29/} {Semantic evaluation for text-to-sql with distilled test suites}.
\newblock In \emph{EMNLP}.

\bibitem[{Zhong et~al.(2017)Zhong, Xiong, and Socher}]{zhong2017seq2sql}
Victor Zhong, Caiming Xiong, and Richard Socher. 2017.
\newblock \href {https://arxiv.org/pdf/1709.00103} {Seq2sql: Generating structured queries from natural language using reinforcement learning}.
\newblock \emph{arXiv preprint}.

\bibitem[{Zhu et~al.(2023)Zhu, Li, Liu, Ma, and Wang}]{zhu2023survey}
Xunyu Zhu, Jian Li, Yong Liu, Can Ma, and Weiping Wang. 2023.
\newblock \href {https://arxiv.org/pdf/2308.07633} {A survey on model compression for large language models}.
\newblock \emph{arXiv preprint}.

\end{thebibliography}
